\documentclass[sigconf]{acmart}

\usepackage{amssymb}

\usepackage{amsmath}
\usepackage{algorithmic}
\usepackage{graphicx}  
\usepackage{float}  
\usepackage{subfigure}  
\usepackage{amsfonts}
\usepackage{multirow, makecell}
\AtBeginDocument{%
  \providecommand\BibTeX{{%
    \normalfont B\kern-0.5em{\scshape i\kern-0.25em b}\kern-0.8em\TeX}}}

\setcopyright{acmcopyright}
\copyrightyear{2023}
\acmYear{2023}
\setcopyright{acmlicensed}\acmConference[HCMA '23]{Proceedings of the 4th
International Workshop on Human-centric Multimedia Analysis}{November 2,
2023}{Ottawa, ON, Canada}
\acmBooktitle{Proceedings of the 4th International Workshop on Human-centric
Multimedia Analysis (HCMA '23), November 2, 2023, Ottawa, ON, Canada}
\acmPrice{15.00}
\acmDOI{10.1145/3606041.3618058}
\acmISBN{979-8-4007-0272-3/23/11}
\acmDOI{10.1145/3606041.3618058}

\begin{document}

\title{Text-based Person Search in Full Images via Semantic-Driven Proposal Generation}

\author{Shizhou Zhang}
\email{szzhang@nwpu.edu.cn}
\affiliation{%
  \institution{Northwestern Polytechnical University}
  \city{Xi'an}
  \state{shaanxi}
  \country{China}
  \postcode{710000}
}

\author{De Cheng}
\authornote{Corresponding author}
\email{dcheng@xidian.edu.cn}
\affiliation{%
  \institution{Xidian University}
  \city{Xi'an}
  \state{shaanxi}
  \country{China}
  \postcode{710000}
}

\author{Wenlong Luo}
\email{luowenlong@mail.nwpu.edu.cn}
\affiliation{%
  \institution{Northwestern Polytechnical University}
  \city{Xi'an}
  \state{shaanxi}
  \country{China}
  \postcode{710000}
}

\author{Yinghui Xing}
\email{xyh_7491@nwpu.edu.cn}
\affiliation{%
  \institution{Northwestern Polytechnical University}
  \city{Xi'an}
  \state{shaanxi}
  \country{China}
  \postcode{710000}
}

\author{Duo Long}
\email{1144849972@qq.com}
\affiliation{%
  \institution{Northwestern Polytechnical University}
  \city{Xi'an}
  \state{shaanxi}
  \country{China}
  \postcode{710000}
}

\author{Hao Li}
\email{2022100658@mail.nwpu.edu.cn}
\affiliation{%
  \institution{Northwestern Polytechnical University}
  \city{Xi'an}
  \state{shaanxi}
  \country{China}
  \postcode{710000}
}

\author{Kai Niu} 
\email{kai.niu@nwpu.edu.cn}
\affiliation{%
  \institution{Northwestern Polytechnical University}
  \city{Xi'an}
  \state{shaanxi}
  \country{China}
  \postcode{710000}
} 

\author{Guoqiang Liang}
\email{gqliang@nwpu.edu.cn}
\affiliation{%
  \institution{Northwestern Polytechnical University}
  \city{Xi'an}
  \state{shaanxi}
  \country{China}
  \postcode{710000}
}

\author{Yanning Zhang}
\email{ynzhang@nwpu.edu.cn}
\affiliation{%
  \institution{Northwestern Polytechnical University}
  \city{Xi'an}
  \state{shaanxi}
  \country{China}
  \postcode{710000}
}

\renewcommand{\shortauthors}{Shizhou Zhang et al.}
\begin{abstract}
  Finding target persons in full scene images with a query of text description has important practical applications in intelligent video surveillance.
  However, different from the real-world scenarios where the bounding boxes are not available, existing text-based person retrieval methods mainly focus on the cross modal matching between the query text descriptions and the gallery of cropped pedestrian images.
  To close the gap, we study the problem of text-based person search in full images by proposing a new end-to-end learning framework which jointly optimize the pedestrian detection, identification and visual-semantic feature embedding tasks.  
  To take full advantage of the query text, the semantic features are leveraged to instruct the Region Proposal Network to pay more attention to the text-described proposals.
  Besides, a cross-scale visual-semantic embedding mechanism is utilized to improve the performance. 
  To validate the proposed method, we collect and annotate two large-scale benchmark datasets based on the widely adopted image-based person search datasets CUHK-SYSU and PRW.
  Comprehensive experiments are conducted on the two datasets and compared with the baseline methods, our method achieves the state-of-the-art performance.
\end{abstract}

\keywords{
text-based Person Search, semantic-driven RPN, cross scale alignment.
}

\maketitle

\section{Introduction}
\begin{figure}[ht] 
	\centering  
	\subfigbottomskip=-2pt 
	\subfigcapskip=-5pt 
	\subfigure[Person ReID]{

		\label{figure_1_a}
		\includegraphics[width=0.94\linewidth,height=0.25\linewidth]{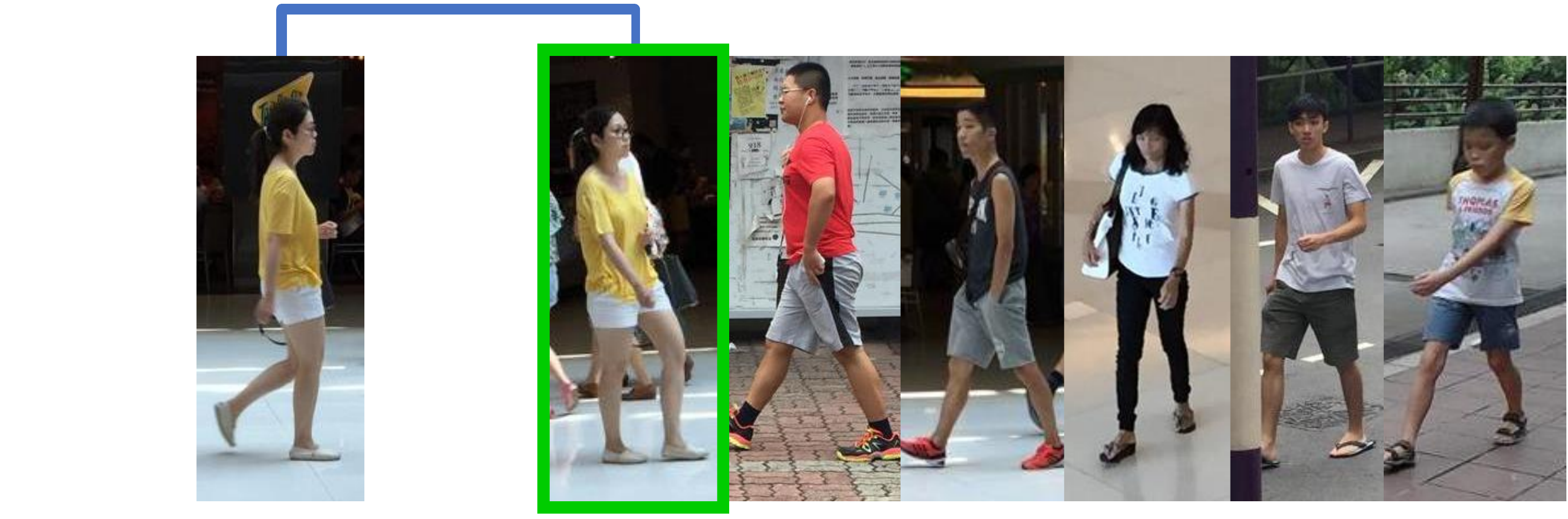}}

	\subfigure[Person Search]{

		\label{figure_1_b}
		\includegraphics[width=0.95\linewidth,height=0.25\linewidth]{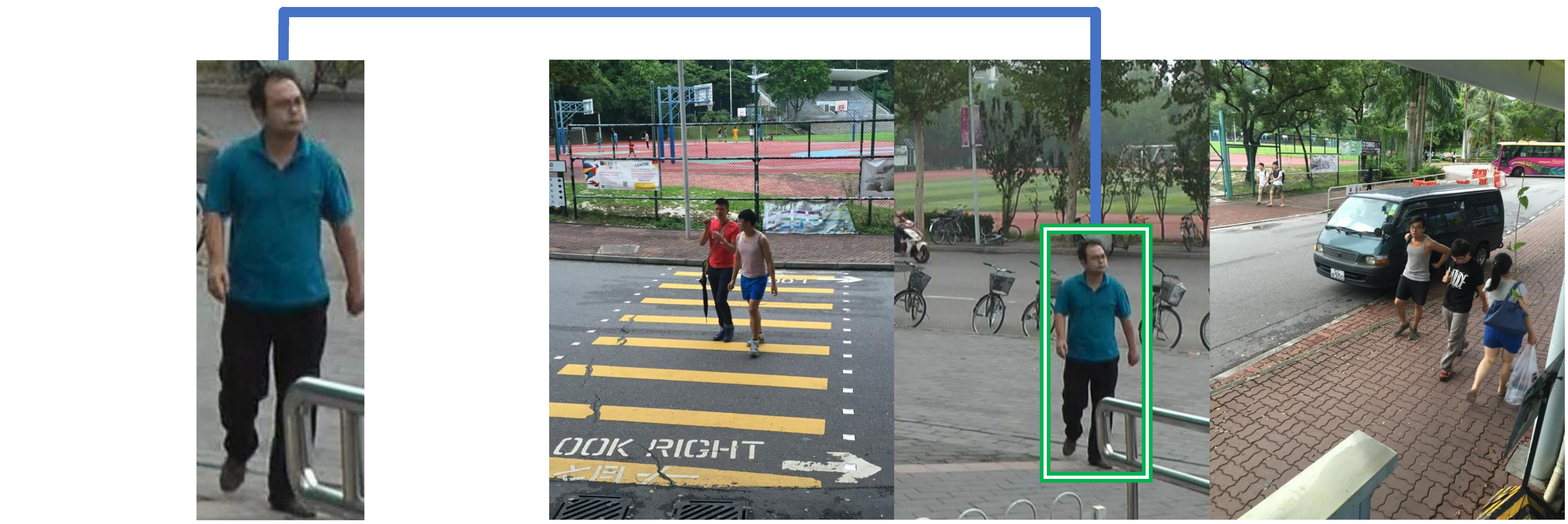}}

	\subfigure[Text-based person retrieval]{

		\label{figure_1_c}
		\includegraphics[width=0.95\linewidth,height=0.25\linewidth]{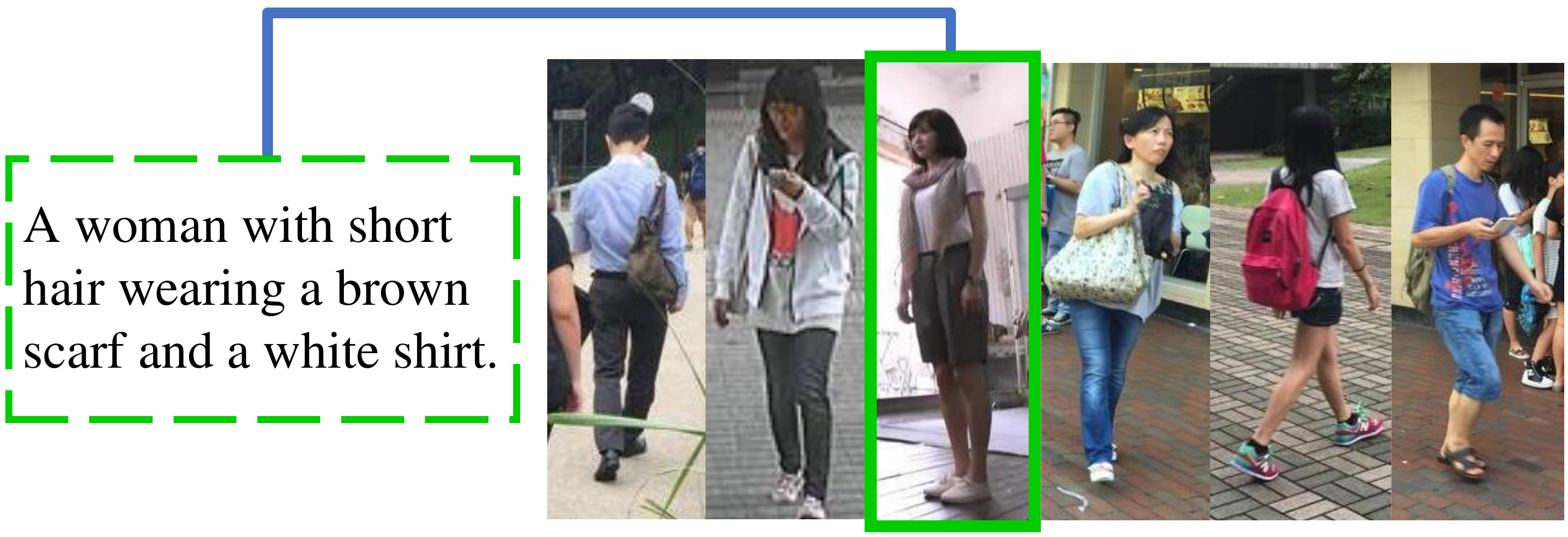}}
	
	\subfigure[Text-based person search]{

		\label{figure_1_d}
		\includegraphics[width=0.95\linewidth,height=0.25\linewidth]{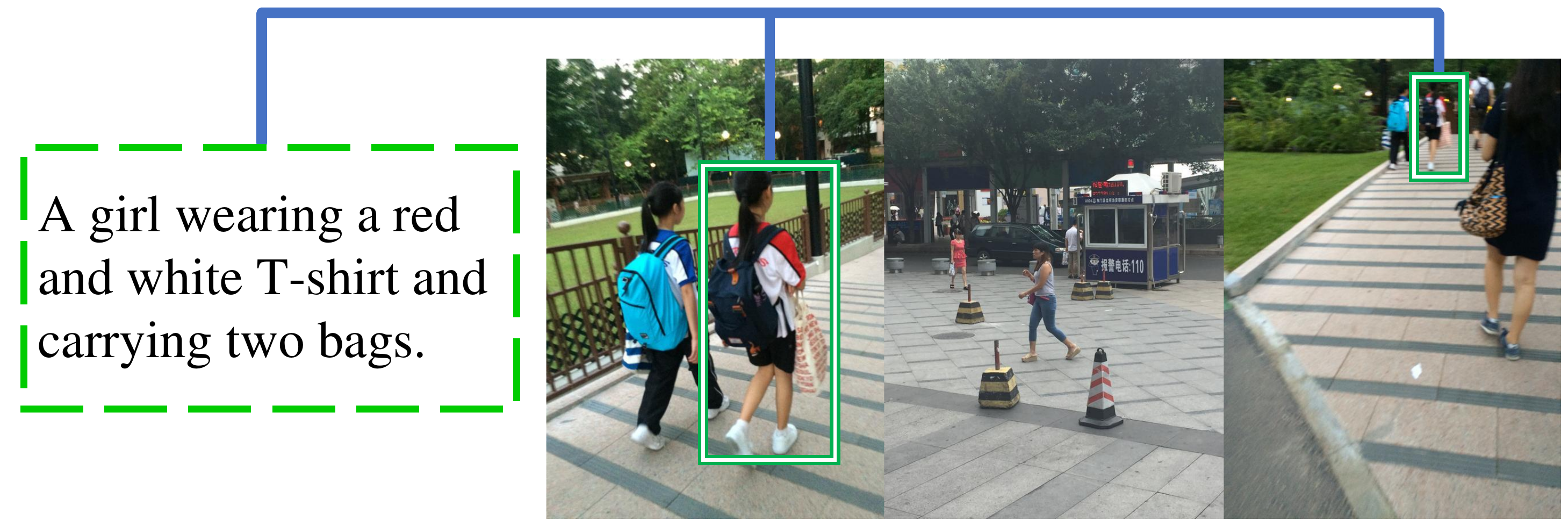}}
	\caption{Comparison of the four tasks. (a) Person ReID. Query: cropped person image. Gallery: cropped person images. (b) Person Search. Query: cropped person image. Gallery: full scene images. (c) Text-based person retrieval. Query: text description. Gallery: cropped person images. (d) Text-based person search. Query: text description. Gallery: full scene images.}
	\label{figure_1}
\end{figure}
Rencently, image-based person reidentification~\cite{ye2021deep,zhang2021person,zhou2018graph,sarfraz2018pose, suh2018part,liu2018pose,Martinel2016reid,li2016null,jiao2021instance} and person search~\cite{xiao2017personsearch,zheng2017PRW,he2018end,zhang2020diverse,wang2019region,lan2018person,munjal2019query, yan2019learning,munjal2019knowledge,islam2020person,zhong2020robust} problems (Figure~\ref{figure_1_a} and ~\ref{figure_1_b}), which aim at matching a specific person with a gallery of cropped pedestrian images or finding a target person in a gallery of full (whole scene) images, have been widely studied in the computer vision community as their great application values in cross camera tracking~\cite{chen2015mirror, zheng2015scalable, zheng2011person, wang2019vehicle, wang2020tcts,zheng2020segmentation,zhai2019fmt,han2019re,dai2020dynamic,chen2020norm,dong2020bi,chen2020a}, criminal investigation, person activity, intention analysis, etc.
In many real-world scenarios, such as finding criminals/suspects, a query image of the target person can not always be easily obtained, while text descriptions given by witnesses are available. 
In such scenarios, it is necessary to develop the techniques for finding a target person with a given query text description.

Although the text-based person retrieval task (Figure~\ref{figure_1_c}), which aims to match a given text query with a gallery of cropped person images, has been explored in recent years~\cite{gao2021contextual, jing2020pose,niu2020improving,chen2018ECCV,zhang2018deep,sarafianos2019adversarial,dong2019person,liu2019deep,ji2018fusion,jing2020cross}.
However, there is still a step distance from the real-world scenarios as the bounding box annotations are unavailable and the query-described person needs to be searched in a gallery of full images.
To close the gap, we study the text-based person search (Figure~\ref{figure_1_d}) problem in this paper.

Note that it is straightforward to breaking down the problem into two independent tasks: person detection and text-based person retrieval. As an off-the-shelf person detector would unavoidably introduce misdetections and misalignments, it could not be optimal when taking the detection results as inputs of the second stage retrieval model. The performance comparison in Section~\ref{sec_expr} also demonstrates the necessity of developing end-to-end method. 

In this paper, we propose a new end-to-end learning framework which integrates person detection, identification and image-text cross-modal matching and jointly optimizes the three tasks together. 
As shown in Figure~\ref{figure_framework}, The detection network follows the Faster-RCNN pipeline, a Region Proposal Network (RPN) for person candidate generation is built on top of a Base-Net which is shared with the identification network. 
Alongside the conventional RPN which aims to output proposals according to the objectness scores, to pay more attention on the text-described proposals and filter out text-irrelevant ones, we propose a novel Semantic-Driven Region Proposal Net (SDRPN) where the RPN features are dynamicly instructed by the semantic feature of the input text description.
After obtaining the features of the proposals, the commonness and positions of the persons are supervised by the detection branch (Det-Net) with a Softmax classification loss and a regression loss,
while the uniqueness of each person IDs are discriminated by enforcing the OIM loss~\cite{xiao2017personsearch} on top of the identification branch (ID-Net).
Furthermore, the BERT language model~\cite{devlin2018bert} is utilized to extract the text features from sentence, sub-sentence and word-levels. 
And the visual features are extracted from global, regional and local scales via splitting and shffuling the proposal feature maps. 
The similarity scores of the proposal-text pairs can be computed with the help of cross attention mechnanism to achieve a cross-scale visual-semantic feature matching.
To validate the proposed method, we collect and annotate two large scale benchmarks for text-based person search based on the widely adopted person search datasets CUHK-SYSU~\cite{xiao2017personsearch} and PRW~\cite{zheng2017PRW}.
To eliminate ambiguity, we name the corresponding datasets as CUHK-SYSU-TBPS and PRW-TBPS, respectively.
As the person datasets already have the annotations of person bounding-boxes and IDs, we merely need to annotate text descriptions for each person bounding-box.  In total,  we collect and annotate 54969 sentences for CUHK-SYSU-TBPS and PRW-TBPS datasets.  The textual descriptions contain abundant noun phrases and various sentence structures. And we give a statistical analysis of the text description in both datasets. Extensive experiments are conducted on these two datasets and the results demonstrate the superiority of our proposed method. Compared with many baseline methods, the proposed method outperforms them by a large margin.

The main contribution of our paper is three-fold and can be summarized as:
\begin{itemize}
\item We make the first attempt to conduct text-based person search in full images, which has more practical application values than text-based person retrieval from cropped pedestrian images.
To support this research direction, two benchmark datasets CUHK-SYSU-TBPS and PRW-TBPS with large scale full images and rich text annotations are collected and annotated.

\item We propose a novel end-to-end learning framework where person detection, identification and image-text embedding tasks are jointly optimized together. 
And it is worth noting that a SDRPN module is devised aiming to care about text description-related person proposals. The proposed SDRPN can boost the final performance by 1.21\% mAP, 1.86\% Rank-1 on CUHK-SYSU-TBPS dataset, and 0.73\% mAP, 1.11\% Rank-1 on PRW-TBPS dataset.

\item We conduct comprehensive experiments on the two datasets and compare our method with many baselines.
The experimental results showed that our method outperforms baseline methods by a large margin and achieves state-of-the-art performances.
\end{itemize}

The rest of the paper is organized as follows: we briefly review related work of our paper in Section~\ref{sec_RW}. Section~\ref{sec_benchmark} gives a statistical analysis of the collected datasets. In Section~\ref{sec_method} we elaborate the proposed framework. The experimental results are reported and analyzed in Section~\ref{sec_expr}. And we conclude the paper in Section~\ref{sec_conc}.

\section{Related Work}\label{sec_RW}
In this section, we briefly review the related works from the following three aspects: 
\subsection{Person search}
Person search is to localize and identify a target person in a gallery of full images other than cropped pedestrian images in ReID task.
Some approaches~\cite{xu2014person, zheng2017PRW, chen2018person} proposed to break down the problem into two separate tasks, pedestrian detection and person re-identification.
Different from the two-stage methods, some works devoted their efforts to propose an end-to-end learning strategy~\cite{xiao2017personsearch,han2019re,munjal2019query} aiming to jointly optimize the detection and re-identification tasks.
Xiao~\cite{xiao2017personsearch} firstly introduced an end-to-end person search network and proposed the Online Instance Matching (OIM) loss function for fast convergence.
Han~\cite{han2019re} proposed to refine the detection bounding boxes supervised by the re-identification training.
Munjal~~\cite{munjal2019query} took full advantage of both the query and gallery images to jointly optimize detection and re-id network.
Additionally, Liu~\cite{liu2017machines} proposed Conv-LSTM based Neural Person Search Machines (NPSM) to perform the target person localization as an  search area iterative shrinkage process. 
Chang~\cite{chang2018rcaa} tranformed the search problem into a conditional decision-making process and trained relational context-aware agents to learn the localization actions via reinforcement learning.

Different from the image based person search whose query is a cropped pedestrian image, in this work, we investigate the text-based person search problem which is much more challenging and able to meet the requirement of the scenarios where query image is not available in many situations.

\subsection{Text-Based Person Retrieval}
Considering that the image query is not always available in real-world scenes, Li~\cite{li2017person} firstly introduced the text-based person retrieval task and collected a benchmark named CUHK-PEDES.
Early methods about text-based person retrieval concentrate on global feature alignment, like~\cite{zheng2017dual,zhang2018deep,sarafianos2019adversarial}, which employed universal feature extraction networks to extract global feature representations for images and descriptions and made efforts to design more proper objective functions for this task.
Such as in~\cite{zhang2018deep}, a cross-modal projection matching (CMPM) loss and a cross-modal projection classification (CMPC) loss were proposed for computing the similarity of image-text pair data.
Meanwhile, there are also several methods~\cite{li2017identity, chen2018improving} employing local feature alignment to provide complementary information for global feature alignment.
For example, Li~\cite{li2017identity} applied the spatial attention, which relates each word with corresponding image regions, to refine the global alignment in the stage-1 training.
Recently, many methods~\cite{chen2018ECCV, gao2020pose, niu2020improving} have applied global and local features of images and text descriptions to realize multi-scale matching and achieved better performance.
Niu~\cite{niu2020improving} proposed a Multi-granularity Image-text Alignment (MIA) module, including global-global, global-local and local-local alignment, and improved the accuracy of retrieval by multi-grained feature alignment between visual representations and textual representations.
Although the multi-scale alignment provide supplement for global feature matching, the alignment for each scale is fixed. 
Gao~\cite{gao2021contextual} realized the need to align visual and textual clues across all scales and proposed cross-scale alignment for text-based person search.

Text-based person retrieval has achieved great performance improvement in recent years, while the task setting still has a gap with the real-world scenarios. Therefore, in this paper, we study text based person search in full images.

\subsection{Variants of Region Proposal Network}
Region proposal network (RPN) is a significant component in series of detection networks, and there are many studies making efforts to improve it in order to generate more accurate or task-relevant proposals.
In order to produce high-quality proposals and improve detection performance, Wang~\cite{wang2019region} proposed Guided Anchoring Region Proposal Network, which learns to guide a sparse anchoring scheme and can be seamlessly integrated into proposal methods and detectors.
Besides, \cite{vu2019cascade} introduces Cascade RPN, which systematically address the limitation of the conventional RPN that heuristically defines the anchors and aligns the features to the anchors for improving the region-proposal quality and detection performance.
To improve the generalization ability of neural networks for few-shot instance segmentation, Fan~\cite{fan2020FGN} proposed attention guided RPN in order to generate class-aware proposals by making full use of the guidance effect from the support set. 

Inspired by the above works, in this paper, we propose a Semantic-Driven Region Proposal Network for text-based person search, which employs semantic information from the query text description to generate semantically similar proposals.

\section{Benchmarks for Text-based Person Search}\label{sec_benchmark}

Since there is no existing datasets for the task of text-based person search, we build new benchmarks for evaluating our method.  
Aiming at this task, the dataset need to include visual information of person bounding-box positions accompanied with person IDs and textual information of language descriptions. Therefore, based on the two widely adopted image-based person search datasets, CUHK-SYSU and PRW, which already contain person bounding-box and ID labels, we match each person box from train set and query set with text descriptions and propose our Text-based Person Search benchmark CUHK-SYSU-TBPS and PRW-TBPS.

For CUHK-SYSU-TBPS, there are 11,206 scene images and 15,080 person boxes with 5532 different IDs in train set, while 2,900 person boxes in query set. And we collect corresponding text descriptions from existing  person retrieval dataset CUHK-PEDES (train\_query\&test\_query), where each person box was labeled with two sentences. 
As for PRW-TBPS, there are 5,704 images and 14,897 boxes, with 483 different IDs in train set and 2,056 boxes in query set.
And text descriptions of all person boxes were annotated, in which the boxes from the train set were labeled with one sentence, and the boxes from the query set were labeled twice independently.
\begin{figure}[ht]
	\centering
	     \subfigure[CUHK-SYSU-TBPS]{
         \label{figure2_a}
	     	\includegraphics[width=0.45 \linewidth]{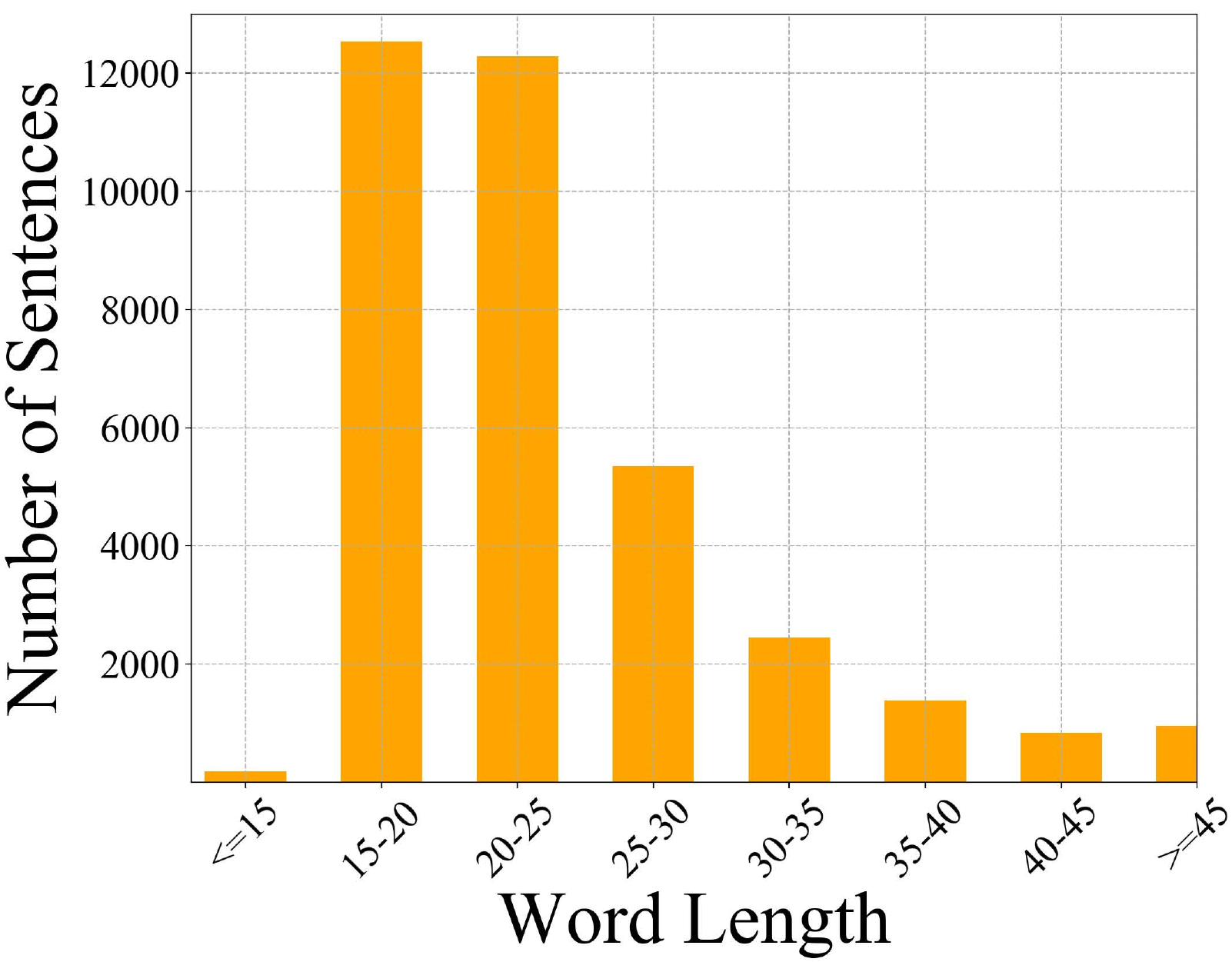}}
	     \subfigure[PRW-TBPS]{
         \label{figure2_b}
	     	\includegraphics[width=0.45 \linewidth]{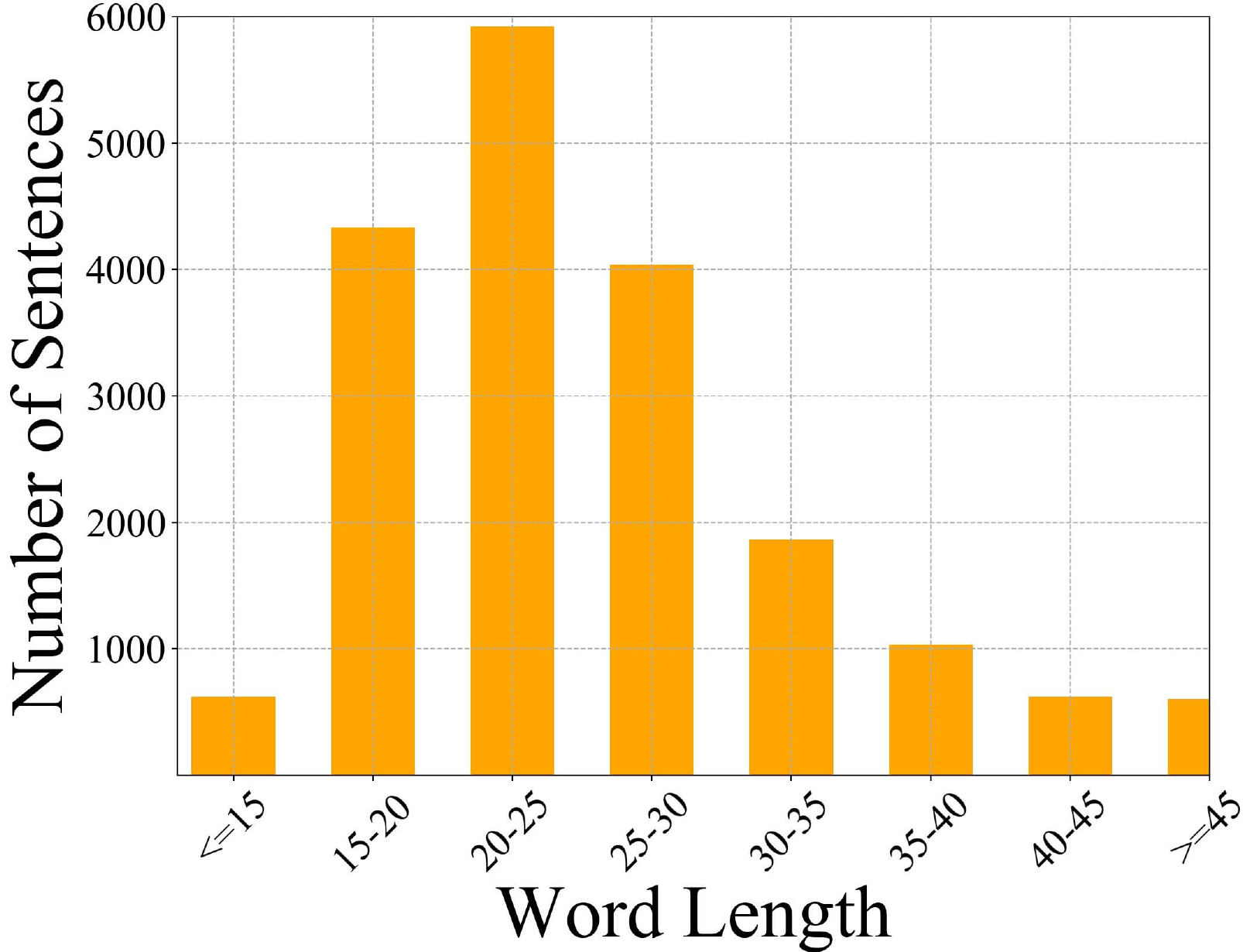}}
	\caption{The  word length distributions on the benchmark datasets CUHK-SYSU-TBPS and PRW-TBPS.}\
	\label{figure_benchmark}
 \vspace{-0.2cm}
\end{figure}
Here, we labeled the training person box once due to the fact that the large amount of repetition of person box share with the same ID, and the average number of each ID occurrence is ten times individuals than that in CUHK-SYSU-TBPS. Therefore, we believe one sentence of each box in PRW-TBPS dataset is capable of providing enough samples for each identity for training.
\begin{figure*}[h!]
	\centering
	\includegraphics[width=\textwidth]{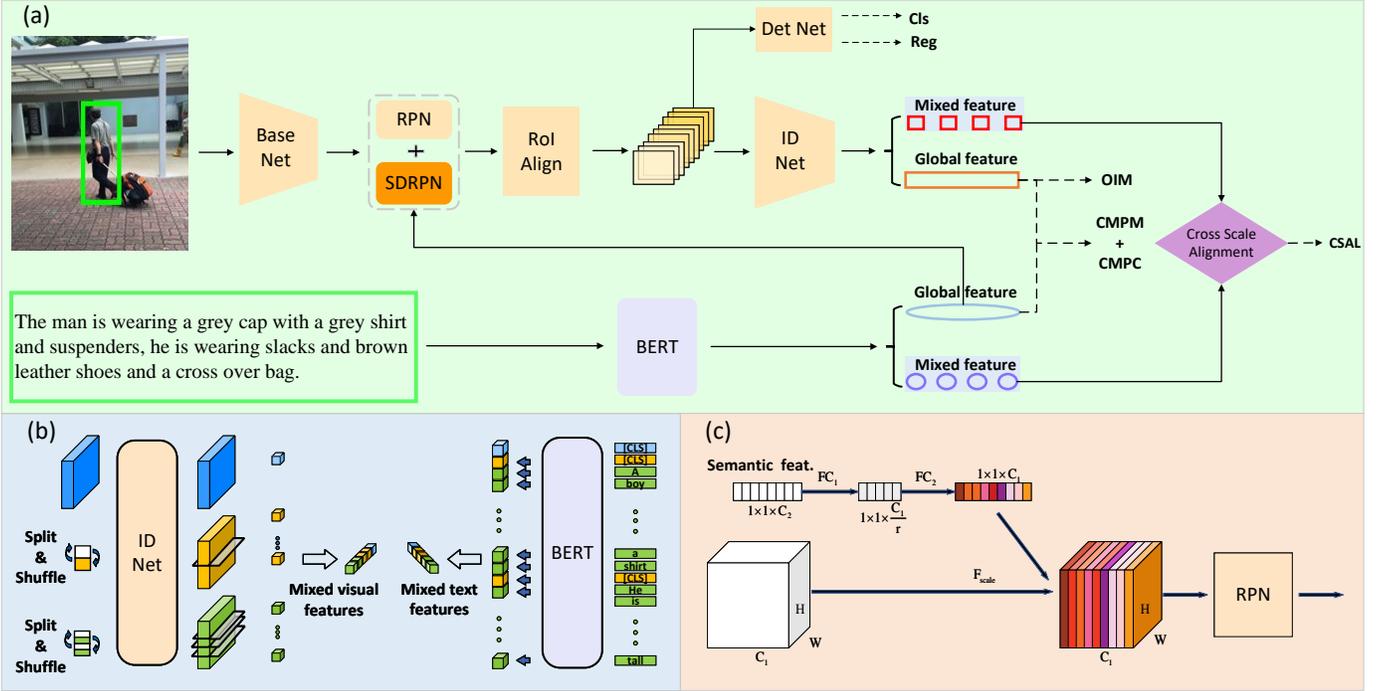}
	\caption{(a) depicts The overall architecture of the proposed learning framework. (b) shows the process of mixed feature extraction. (c) exhibits the SDRPN.}
	\label{figure_framework}
\end{figure*}

The text descriptions of the datasets not only focus on person appearances, including clothes and body shape, but also pay attention to person actions, gestures and other details. 
To some extent, vocabulary and sentence length are vital indicators to evaluate the capacity of the dataset. In total, there are 1,318,445 words and 5,934 unique words in the datasets. As Figure~\ref{figure_benchmark} shows, most sentences have 15 to 45 words in length, and the average word lengths of the datasets are 23.9 and 24.96 words respectively, which is much longer compared with other image-caption datasets like MS-COCO~\cite{lin2014microsoft} (5.18 words in average) and Visual Genome~\cite{krishna2017visual} (10.45 words in average). 

\section{Method}\label{sec_method}
In this section, we introduce the proposed end-to-end learning method as illustrated in Figure~\ref{figure_framework}.
Firstly, we briefly introduce the overview of the framework (Figure~\ref{figure_framework} (a)).
Then two major components, namely Semantic-Driven RPN module and 
Proposal-Text Embedding module are elaborated.
Finally, we give the total loss function of training the proposed method.

\subsection{Overview}
Our goal is to search the target person from full images with a query of text description. 
To be exact, we decompose the task to three sub-tasks, namely pedestrian detection, identification and image-text cross-modal feature embedding/matching.
As shown in Figure~\ref{figure_framework} (a), the whole framework takes full images with text descriptions of labeled persons as input when training. The framework includes two paths, namely image-path and text-path.

For the image path, we follow the structure of one-stage person search method and add a multi-task head for localization, detection, and identification on top of the convolutional features of Faster-RCNN.
We first exploit the ResNet-50 as backbone and split it into two parts as Base-Net (Conv1 to Conv4-3) and ID-Net (Conv4-4 to Conv5).

As for the text-path, the semantic feature of the query text is encoded by a BERT language model~\cite{devlin2018bert}.
Then both the visual feature and semantic feature from the two paths are fed into proposal-text embedding module. 

\subsection{Semantic-Driven RPN}
The goal of our end-to-end text-based person search task is to ``detect'' the right person matching with the text query. We propose SDRPN to filter out the irrelevant ones. Traditional RPN generates class-agnostic proposals for generic object detection. 
To take full advantage of the text query which contains full semantic information,
we devise a Semantic-Driven RPN which leverages the semantic features from the text query to instruct the proposal generation process, aiming at paying more attention to those semanticly more similar candidates with the text description.
Sepecifically, inspired by the SENet~\cite{hu2018squeeze}, the semantic features are utilized to re-weight the Base-Net feature maps. 
As illustrated in Figure~\ref{figure_framework} (c), SDRPN includes a channel-wise attention mechanism to guide a standard RPN, generating the proposal boxes from the re-weighted image features.

In more detail, we use the semantic feature extracted from a BERT language model and unsqueeze it to $1\times1\times C_{2}$.
The resulted feature is denoted as $\textbf{z}$ and then we apply two fully connected layers $FC_{1}$ and $FC_{2}$ to squeeze and expand the feature $\textbf{z}$ , from $C_{2}$ to $C_{1}/r$ then to $C_{1}$, to emphasize important signal correlations. 
Based upon the sigmoid activation $\sigma$, the resulted excitation $\textbf{s}$ is computed as follows.
\begin{equation}
	\textbf{s}=FC_{2}(FC_{1}(\textbf{z}))=\sigma(\textbf{W}_{2} \bf{\delta} (\textbf{W}_{1}\textbf{z})).
\end{equation}
Then the excitation $\textbf{s}$ is applied to the BaseNet feature maps $\textbf{X}$ from a gallery image through channel-wise multiplication as follows:
\begin{equation}
	\hat{\textbf{X}} = F_{scale}(\textbf{X},\textbf{s}) = \textbf{s} \odot \textbf{X} 
\end{equation}
Note that SDRPN extracts proposals featuring at a text-similarity score and RPN pursues the standard objectness score, 
Therefore, as shown in Figure~\ref{figure_framework} (a), we use SDRPN in parallel with RPN when generating proposals by summing up the scores of corresponding anchor boxes to obtain better performance.

\subsection{Proposal-Text Embedding}

The proposal-text feature embedding module aims to learn a common space for both the visual and text modality. To improve the performance, a cross-scale alignment scheme is borrowed in the embedding process.

 \textbf{Multi-Scale Visual Feature Extraction.}
 In visual path, proposal features will be represented in three scales from coarse to fine, namely as global-scale, region-scale and local-scale.
As illustrated in Figure~\ref{figure_framework} (b), we take the output of ID-Net as the global-scale representations of the proposals.
Further, to better focus on local features and reduce the influence of large receptive field of CNN, we do the split and shuffle operation on the RoI-Aligned proposal features, which equally partitions the feature map into several horizontal stripes, and these set of the partitioned stripes are randomly shuffled and re-concatenated.
The re-concatenated feature maps then are passed through the ID-Net.
After that, the output feature map of the region-scale branch is horizontally partitioned into $n$ stripes, each of which is further encoded as a region-scale feature corresponding to a certain region. 
Finally, a finer partition scheme is used to produce the local-scale features,

 \textbf{Multi-Scale Semantic Feature Extraction.}
As for the text path, we use the BERT language model to extract the semantic representation from three levels, namely sentence-level, sub sentence-level and word-level (Figure~\ref{figure_framework} (b)). 
We use the final hidden state of token [CLS], which is added at the beginning of the sentence, as the representation of the whole sentence.
For the sub sentence-level, sentences are separated by commas resulting in several shorter sub-sentences. 
And we attach the [CLS] token to the beginning of each sub-sentence, whose final hidden state is treated as the representation of each sub-sentence. 
While as for the word-level, each final hidden state of word is considered as the word-level representation.

\begin{table*}
  \centering
  \caption{Performance comparison of the baseline methods and the proposed method on  CUHK-SYSU-TBPS and PRW-TBPS.}
  \label{table_resultscomp}
  \begin{tabular}{ccccccccc}
    \toprule
    \multirow{2}{*}{Method} & \multicolumn {4}{c}{CUHK-SYSU-TBPS} & \multicolumn {4}{c}{PRW-TBPS}\\
    \cmidrule(r){2-5} \cmidrule(r){6-9}
    & mAP(\%)  & top-1(\%) & top-5(\%) & top-10(\%)
    & mAP(\%)  & top-1(\%) & top-5(\%) & top-10(\%)\\
    \midrule
     OIM+BiLSTM &23.74	&17.41	&38.48 &49.21	&4.58	&6.66	&16.33	&22.99\\
     NAE+BiLSTM &23.48	&16.62	&38.45	&49.66	&5.20	&7.54	&17.21	&24.11\\
     BSL+BiLSTM &26.91	&20.97	&42.31	&52.31	&3.60	&6.42	&15.41	&22.46\\
     OIM+BERT &43.39	&36.59	&62.03	&72.66	&8.52	&14.44	&30.68	&39.77\\
     NAE+BERT &45.70	&39.14	&64.62	&74.34	&9.20	&14.44	&31.55	&39.91\\
     BSL+BERT &48.39	&40.83	&67.52	&76.86	&10.70	&16.82	&34.86	&45.36\\
    \midrule
     Ours &\textbf{50.36}	&\textbf{49.34}	&\textbf{74.48}	&\textbf{82.14}	&\textbf{11.93}	&\textbf{21.63}	 &\textbf{42.54}	&\textbf{52.99}\\
  \bottomrule
\end{tabular}
\end{table*}

\textbf{Proposal-Text cross scale alignment.} 
After proposal and text feature extraction, we obtain a set of three-scale visual and semantic features.
We concatenate them to get the mixed visual features
$\textbf{I}=\{\textbf{I}_1,\textbf{I}_2,... ,\textbf{I}_m \}$ and mixed textual features $\textbf{T} = \{\textbf{T}_1,\textbf{T}_2,... ,\textbf{T}_n\}$, where m and n corresponds to the $m_{th}$ and $n_{th}$ part.
To get the cross attended features, fully connected layers are used to map the mixed visual features $\textbf{I}$ to visual queries, keys and values, denoted by $\textbf{Q}$, $\textbf{K}$ and $\textbf{V}$ with weight matrix $\textbf{W}_\textbf{q}$,$\textbf{W}_\textbf{k}$ and $\textbf{W}_\textbf{v}$, respectively. 
And the mixed semantic features $\textbf{T}$ are mapped to semantic queries, keys and values, denoted by $\mathcal{Q}$, $\mathcal{K}$ and $\mathcal{V}$.


Firstly, the attended semantic feature $\textbf{A}$ can be computed from the view of text-to-image attention mechanism as,
\begin{equation}  
 \textbf{A}= norm(\textbf{Q} \mathcal{K}^{T}) \odot \mathcal{V}
\end{equation}

Then we can obtain the relevance between the visual value and its corresponding semantic context by calculating the cosine similarity between $\textbf{V}$ and $\textbf{A}$,
\begin{equation}
\textbf{R}=\cos (\textbf{V} \odot \textbf{A})
\end{equation}
where $\textbf{R}$ denotes the relevance scores.
The similarity of text-to-image pair $S$ is then computed by averaging all components of $\textbf{R}$.
Meanwhile, by alternating the semantic keys as queries and visual queries as keys respectively, and following the above procedure, the similarity of image-to-text pair 
denoted by $S^{\prime}$ can be computed.

Then, assuming that a mini-batch of $B$ person boxes and captions are given, and all image-to-text pairs are constructed as $ \{(\textbf{I}_{i},\textbf{T}_{j}),y_{i,j} \}_{i=1,j=1}^{B}$. 
Note that $y_{i,j} =1$ if $(\textbf{I}_{i},\textbf{T}_{j})$ is a matched pair, otherwise $y_{i,j} = 0$. 
To maximize similarities between the matched pairs and push away the unmatched pairs, KL divergence is enforced to diminish the modality discrepancy.
Considering that the normalized similarities $\overline{\textbf{S}}$ can be treated as the predicted matching probability and the normalized label vector $\overline{\textbf{y}}$ can denote ground-truth label distribution.
\begin{align}\label{eqn_q}
	L_{I}  = D_{KL} ( \overline{\textbf{S}},\overline{\textbf{y}}) \, \nonumber \\
	L_{T}= D_{KL}(\overline{\textbf{S}^{\prime}},\overline{\textbf{y}^{\prime}})
\end{align} 

Finally, the Cross-Scale Alignment Loss (CSAL) is calculated by,
\begin{equation}
L_{CSAL}  =  L_{I}+L_{T}
\end{equation}

\textbf{Global matching.} 
Besides the cross scale alignment with mixed features, we additionally use CMPM loss and CMPC loss~\cite{zhang2018deep} to supervise the cross modal matching of the global-scale features. 
CMPM loss computes the matching probability of the proposal-text pair
and CMPC is a variant of norm-softmax classification loss.
We refer readers to ~\cite{zhang2018deep} for more details about these two losses. 



\subsection{Overall Loss Function}
The whole framework is trained via an end-to-end strategy and pursue the joint optimization of all the loss functions for each task. 
More specifically, the sub-network for person detection is supervised with a classification loss ($L_{cls}$), a regression loss ($L_{reg}$),  a RPN objectness loss ($L_{rpn\_cls}$),  and a RPN box regression loss ($L_{rpn\_reg}$). 
While for the supervision of the identification network,  the adopted loss function is the OIM loss ($L_{oim}$).  
To learn a common feature space for proposals and text descriptions, we adopt CMPM Loss ($L_{cmpm}$), CMPC Loss ($L_{cmpc}$), and cross-scale alignment loss ($L_{csal}$). Therefore, the overall loss function is formulated as:
\begin{equation}
\begin{split}
	L = \lambda_{1}L_{rpn\_cls}+\lambda_{2}L_{rpn\_reg}+\lambda_{3}L_{cls}+\lambda_{4}L_{reg}\\	        			+\lambda_{5}L_{oim}+\lambda_{6}L_{cmpm}+\lambda_{7}L_{cmpc}+\lambda_{8}L_{csal}
\end{split}
\end{equation} 
where $\lambda_1-\lambda_8$ are responsible for the relative loss importance.

\subsection{Inference}
During the inference time, the global text feature and global visual features are extracted to represent the textual query and candidate proposals. 
The text query features are extracted from the fine-tuned BERT model, while quantities of proposal features are obtained from the trained visual module of joint-optimized model after inputting corresponding gallery images. 
Then, we compute the cosine similarity between the query feature and proposal features to sort and rank the candidate bounding boxes.
  
\section{Experiment}\label{sec_expr}

In this section, we report and analyze the experimental results on the collected datasets.
Firstly, we describe the details of datasets and evaluation protocols as well as the implementation details.
To verify the effectiveness of the proposed end-to-end approach, we investigate several two-stage solutions as baseline methods. 
In addition, we conduct ablation studies to analyze the influence of each component in our proposed method.
Finally, both quantitative and qualitative results are exhibited.

\subsection{Datasets and Protocol}

The collected datasets are built upon the two existing image-based person search datasets CUHK-SYSU and PRW.  
On CUHK-SYSU-TBPS, The testing set includes 2,900 query descriptions and 6,978 gallery images. 
For each query, different gallery sizes are set to assess the scaling ability of different models. 
We use the gallery size of 100 by default. 
As for PRW-TBPS, the testing set contains 2,057 query persons and each of them are to be searched in a gallery with 6,112 images.

To measure the performance of text-based person search task,  the widely adopted mean Average Precision (mAP) and Cumulative Matching Characteristics (CMC top-K) are used as standard metrics. 
However, different from the conventional retrieval tasks, a candidate in the ranking list would only be considered correct when its IoU with the ground truth bounding box is greater than 0.5.


\begin{table*}[ht]
    \centering
    \small
	\caption{Performance comparison of different components in our method.}
	\label{tab:freq}
	\begin{tabular}{ccc|cc|cc}
		\toprule
		\multirow{2}{*}{BERT} & \multirow{2}{*}{Cross-scale Alignment}  & \multirow{2}{*}{SDRPN} & \multicolumn {2}{|c|}{CUHK-SYSU-TBPS} & \multicolumn {2}{c}{PRW-TBPS} \\
        \cmidrule(r){4-5} \cmidrule(r){6-7}
		 & & & mAP(\%)  & top-1(\%) 
		 & mAP(\%)  & top-1(\%)\\ 
		\midrule
		$\surd$ &$\times$ &$\times$ &48.77 &45.86 &10.48&19.40\\
		$\surd$ &$\surd$ &$\times$ &49.15 &47.48 &11.20 &20.52\\
		$\surd$ &$\surd$ &$\surd$ &\textbf{50.36} & \textbf{49.34} &\textbf{11.93} &\textbf{21.63}\\
		\bottomrule
	\end{tabular}
\end{table*}

\begin{table*}
    \centering
	\caption{Performance comparison of two-stage methods and ours on PRW-TBPS dataset.}
	\label{tab:freq}
	\begin{tabular}{c|ccccc}
		\toprule
		\multirow{1}{*}{Model} & \multirow{1}{*}{mAP}  & \multirow{1}{*}{top-1\%} & \multicolumn {1}{c}{top-5\%} & \multicolumn {1}{c}{top-10\%} & \multicolumn {1}{c}{inference time}\\
		\midrule
        DETR+``CMPM+CMPC'' & $1.01$ & $1.65$  & $6.91$ & $15.47$ & $0.19s$\\
        Faster-RCNN+``CMPM+CMPC'' & $1.03$ & $2.14$  & $8.17$ & $12.84$ & $0.21s$\\
        Ours & $11.93$ & $21.63$  & $42.54$ & $52.99$ & $0.98s$\\
		\bottomrule
	\end{tabular}
\label{tab:efficiency}
\end{table*}

\subsection{Implementation Details}

We take ImageNet-pretrained ResNet-50 to initialize parameters of Base-Net and ID-Net. 
When obtaining the mixed visual features, the region branch splits the proposal feature map into two stripes equally and the local branch splits the feature map into three stripes equally. The dimension of the visual features at different scales is set to 768-d. 
For the text semantic feature extraction module, we use BERT-Base-Uncased model as the backbone, which is pretrained on a large corpus including Book Corpus and Wikipedia. 
The dimension of different scale textual features is also set to 768.

In RPN and SDRPN, we adopt the same anchors and adjust the anchor sizes to the objects in the dataset. 
We adopt scales \{8, 16, 32\} and aspect ratios \{1, 2\}. 
Non maximum Suppression (NMS) with a threshold of 0.7 is used to filter out redundant boxes and 2000 bounding boxes are left from original 12,000 bounding boxes after NMS. 
Then, we select top 4 proposals for each person identity from 2000 RoIs based on objectiveness score, and meanwhile the IoU of the selected proposals have to be bigger than a threshold of 0.5. 
During inference, we keep 300 boxes from 6000 bounding boxes and sent them into detection branch. 
In SDRPN, the reduction ratio $r$ is set to 16. 
The loss function of SDRPN and RPN are both cross-entropy loss. Note that in SDRPN, anchor boxes that overlap with the text-relevant persons are marked as positives. 
While in the standard RPN, all persons in the image are positive samples. 

The batch size is set to 4, and we use horizontally flipping as data augmentation. 
The model contains three groups of parameters, namely detection, identification and projection parameters. 
The detection parameters are optimized with SGD optimizer with momentum of 0.9, and identification and projection parameters adopt Adam optimizer. 
The learning rate of three groups parameters are set to 0.0001, 0.001, 0.0001 respectively, and the model is trained for 12 epochs in total.
The hyper-parameters of each loss function are set to 1, except the one for CSAL loss which is set to 0.1. 

\subsection{Compared Methods}

Since there is no existing method specifically designed for text-based person search, we explore typical methods of related tasks and  split the task into two parts, detection and text-image alignment, which are combined together as two-stage method to compared with our proposed one stage model. 

Specifically, we take fully trained person search model to extract visual features of labeled person image. 
Also we use language model to extract textual features of language description. 
The distances between visual feature and text feature are measured under the supervision of CMPM and CMPC loss. 
During inference, the similarity between the query text and the detected person bounding boxes is calculated based on their embedded features. 

The chosen person search method contains OIM~\cite{xiao2017personsearch}, NAE~\cite{chen2020norm}, and BSL, which are all based on Faster-RCNN while the model architectures are different.
In OIM, the box regression and region classification losses remain the same as in Faster-RCNN, with an additional identity classification loss as supervision.
In contrast, NAE removes the original region classification branch and uses the embedding norm as the binary person/background classification confidence. 
BSL is the network used in our framework which is also evaluated as an image-based person search method. 
Different with OIM and NAE, BSL uses one convolution layer instead of identification net for detection branch, 
meanwhile the output feature of identification net is directly encoded as final feature vector for matching without further projection to reduce the feature dimension.

As for language model, BiLSTM and BERT are both used as text  feature extractors. 
Notebly, the number of hidden units of BiLSTM is set to 2048 when matching visual features extracted by BSL, otherwise the hidden units number is 256. 
While, for BERT, we use $1\times 1$ convolution to adjust the shape of text features. 
All BiLSTM networks are trained for 150 epochs and BERT is trained with 50 epochs.

\begin{figure}[t]
	     \subfigure [Top-1 result ]{
	     	\includegraphics[width=0.45 \linewidth]{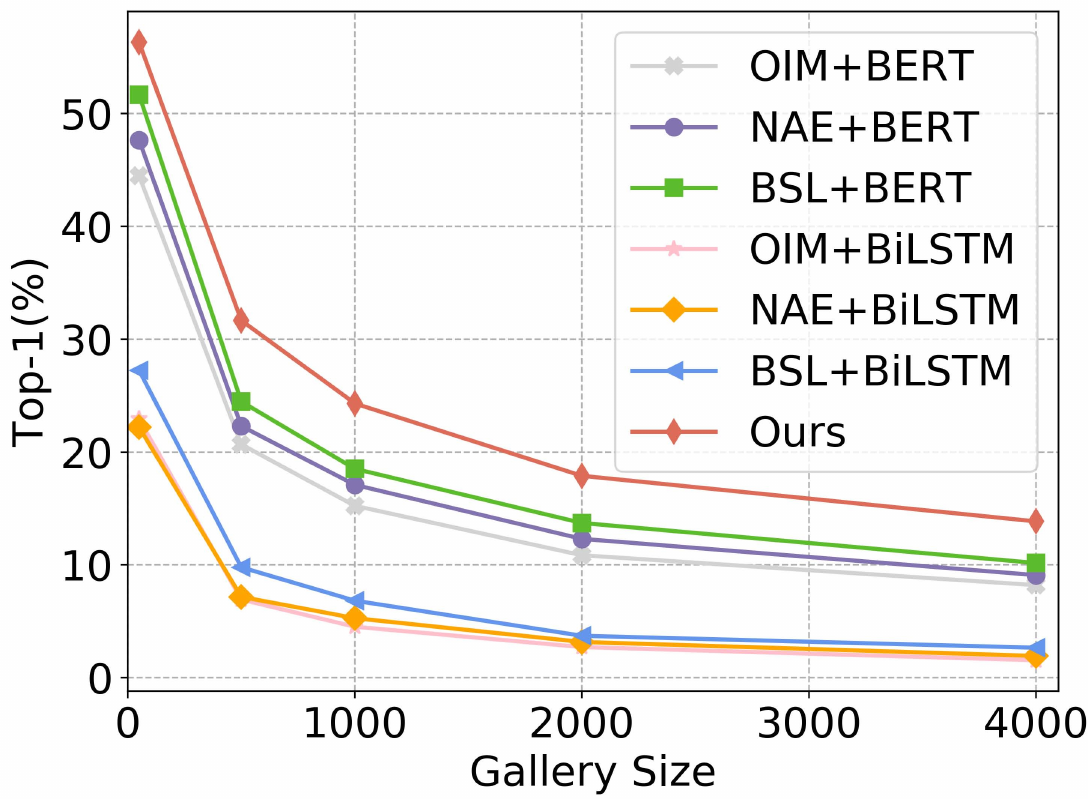}}
	     \subfigure[mAP result]{
	     	\includegraphics[width=0.45 \linewidth]{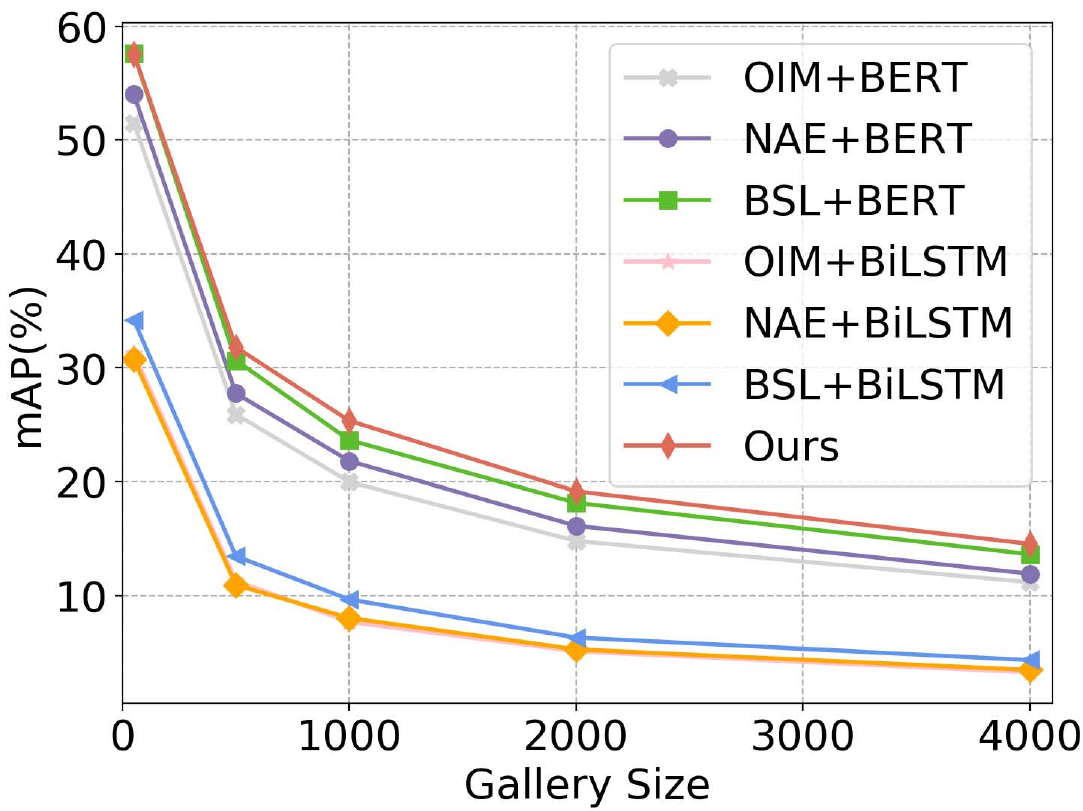}}
	\caption{Results comparison with different gallery size of CUHK-SYSU-TBPS }\
	\label{figure_gallerysize}
\end{figure}

\subsection{Quantitative and Qualitative Results}

\begin{figure}[ht]
	\centering
	\includegraphics[width=0.95\linewidth, height=0.55\linewidth]{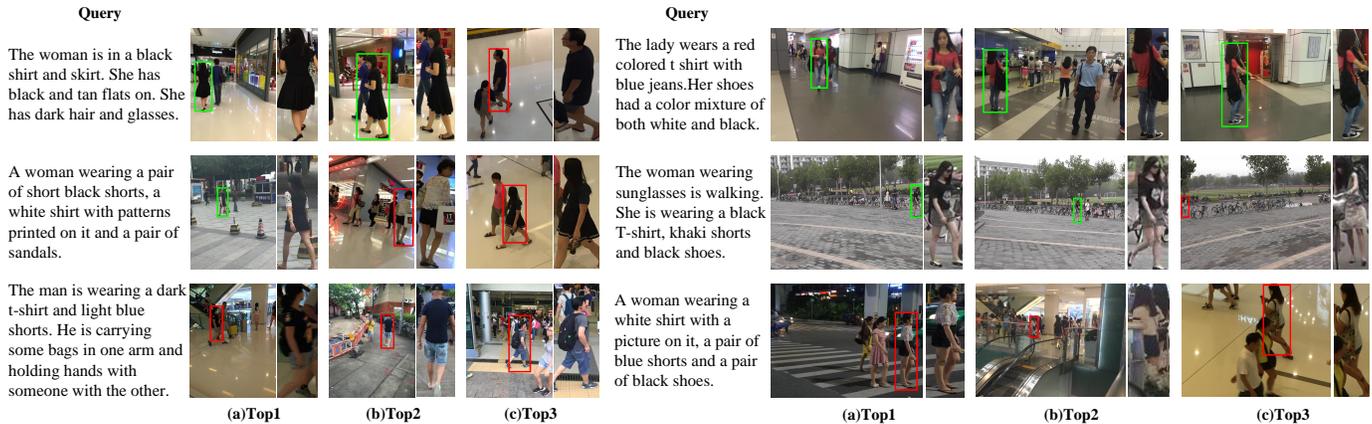}
	\caption{Examples of text-based person search.}
	\label{figure_exampleresults}
    \vspace{-0.6cm}
\end{figure}

	

\textbf{Comparison with baseline methods.}
Table~\ref{table_resultscomp} shows the results of our proposed framework and the compared two stage methods on both the datasets.
On CUHK-SYSU-TBPS dataset, our method acheived 49.34\% Rank-1 accuracy and 50.36\% mAP, which is +8.51\% Rank-1 and +1.97\% mAP better than the superior compared method BSL+BERT.
On the more challenging PRW-TBPS dataset, our method acheived 21.63\% Rank-1 accuracy and 11.93\% mAP, which is +4.81\% Rank-1 and +1.23\% mAP better than BSL+BERT method.
As can be seen, our approach achieves state-of-the-art performances in terms of both mAP and CMC top-1 to 10 accuracies.

It can also be clearly seen from Table~\ref{table_resultscomp} that (1) When using BERT as the text feature extraction model, it brings significant improvement for our task compared with BiLSTM on both datasets. It indicates that BERT is more capable of encoding complex text descriptions into semantic feature vectors for joint alignment with visual features in a certain way. (2) BSL architecture is more suitable for the task compared with OIM and NAE, as about 1\%-3\% improvement can be obtained in terms of both CMC top-1 accuracy and mAP on both datasets. We infer that the usage of separate Det-Net and ID-Net for detection and identification, is better for person search model to obtain more accurate location of detection and more discriminative  visual features. (3) The proposed end-to-end solution has clearly advantages as it can beat all the two-stage counterparts.

\textbf{Results over varying gallery sizes on CUHK-SYSU-TBPS.}
As shown in Figure~\ref{figure_gallerysize}, when the gallery size of CUHK-SYSU-TBPS is adjusted from 50, 500, 1000, 2000 to 4000, all of the methods degraded the performance while our method exhibits the consistent advantages compared with others.


\textbf{Component analysis.}
We analyze three major components of our method, namely BERT, Cross-scale Alignment and SDRPN, by observing the performance improvement when progressively adding each component. 
The results are reported in Table~\ref{tab:freq}.
The first row of Table~\ref{tab:freq} is a baseline one stage model which adopts BERT to extract text features with a standard RPN. CMPC and CMPM loss are used for training the model.
Note that even the baseline one stage model outperforms the best two stage model. 
Then, we introduce cross-scale alignment for extracted mixed features and add CSAL loss for joint text-image embedding, which brings +1.62\% and +1.12\% performance improvement in terms of CMC top-1 accuracy on the two datasets.
Based upon that, SDRPN when combined together with standard RPN as aforementioned improves the CMC top-1 accuracy by additional +1.86\% and +1.11\% on the two datasets, respectively.

We also campare the  the two-stage methods , i.e. detection + text-based re-id with our end-to-end text-based person search method. For detection, we select two classical detection methods, Faster-RCNN and  DeTR, to serve as the detection method. And we adopt a text-based re-id method ``CMPC+CMPM''\cite{zhang2018deep} as its complexity is more or less comparable with our method.
The experimental results are shown in Table~\ref{tab:efficiency}.Our end-to-end method is significantly better than the two-stage methods in performance, which outperform 10.8\% in mAP and it shows the necessity of developing end-to-end person search methods.

To further verified whether the global feature and the mixed feature should be aligned separately using different losses, we replace CSAL with CMPM+CMPC to conduct the experiment, and we found that the final results is a little inferior as in Table~\ref{tab:exchange}, the origin setting is outperform the replacing setting about 2\%.

To find the best hyper-parameter for the CSAL, We did the hyper-parameter experiments on ${\lambda}_8$. The specific experimental results are shown in Table~\ref{tab:coefficient}. The experiments shows that higher CSAL coefficient takes negative effect of the final results, which may be mainly due to the fact that the magnitude of CSAL loss itself is greater than the other two losses.

\begin{table} [ht]
    \centering
    \renewcommand\thetable{\uppercase\expandafter{\romannumeral4}}
	\caption{Replacing the CSAL loss with CMPM.}
	\label{tab:freq}
	\begin{tabular}{c|cccc}
		\toprule
		\multirow{1}{*}{Model} & \multirow{1}{*}{mAP}  & \multirow{1}{*}{top-1\%} & \multicolumn {1}{c}{top-5\%} & \multicolumn {1}{c}{top-10\%}\\
		\midrule
        W/ Replacing & $9.93$ & $18.77$  & $39.67$ & $50.95$\\
        W/O Replacing & $11.93$ & $21.63$  & $42.54$ & $52.99$\\
		\bottomrule
	\end{tabular}
\label{tab:exchange}
\end{table}

\begin{table} [ht]
    \centering
    \small
    \renewcommand\thetable{\uppercase\expandafter{\romannumeral5}}
	\caption{The experiments to tuning the coefficient $\lambda^{}{\_}8$.}
	\label{tab:freq}
	\begin{tabular}{c|cccc}
		\toprule
		\multirow{1}{*}{Model} & \multirow{1}{*}{mAP}  & \multirow{1}{*}{top-1\%} & \multicolumn {1}{c}{top-5\%} & \multicolumn {1}{c}{top-10\%}\\
		\midrule
        0.1 & $11.93$ & $21.63$  & $42.54$ & $52.99$\\
        0.2 & $10.74$ & $20.61$  & $38.89$ & $48.86$\\
        0.5 & $11.66$ & $22.22$  & $43.76$ & $53.48$\\
        1 & $10.38$ & $21.29$  & $38.23$ & $49.64$\\
		\bottomrule
	\end{tabular}
\label{tab:coefficient}
\end{table}

\textbf{Qualitative results.}
Figure~\ref{figure_exampleresults} illustrates some text-based person search results. 
The boxes with green lines represent correct search results, while the boxes with red lines denote failure results. 
The top 2 rows demonstrate successful cases where correct person boxes are within the top-3 retrieved full images. 
From these successful cases, we can observe that our method can spot the target person occoured with different angle and background in full scene images.
Even though in some cases, like the second case of the middle row in Figure~\ref{figure_exampleresults}, the size of person box is relative small compared to the full scene images, it can also be correctly searched through a text description by our model. 
Meanwhile, in failure cases, some search results have some characteristics that partially fits the query description, such as the bottom-left case in Figure~\ref{figure_exampleresults}, the first two persons both wear black T-shirt and the third man carries a black backpack. And they all wear blue pants, which are very close to part of the query description.

\section{Conclusion}\label{sec_conc}
In this paper, we investigate the problem of text-based person search in full scene images to meet the real-world scenarios where both the query image and the bounding boxes are not available.
Specifically, instead of a straightforward two-stage method, we proposed a new end-to-end learning framework which integrated the pedestrian detection, person identification and image-text cross-modal feature embedding tasks together and jointly optimize them to achieve better performance.
To take full advantage of the query text description, we devise a Semantic-Driven Region Proposal Network where the proposal generation process is instructed to pay attention to those candidates which are more similar with the semantic features of the text description.
Furthermore, a cross-scale visual-semantic feature matching mechanism is introduced to improve the final searching results.
To validate the proposed approach, we collect and annotate two large scale text-based person search benchmark datasets named as CUHK-SYSU-TBPS and PRW-TBPS which are built on top of the widely adopted image-based person search datasets CUHK-SYSU and PRW, respectively.
We conduct extensive experiments and the experimental results on the two datasets demonstrated that our proposed method achieved state-of-the-art performance compared with many classical baseline methods.

\bibliographystyle{ACM-Reference-Format}
\bibliography{reference}


\begin{thebibliography}{57}


\ifx \showCODEN    \undefined \def \showCODEN     #1{\unskip}     \fi
\ifx \showDOI      \undefined \def \showDOI       #1{#1}\fi
\ifx \showISBNx    \undefined \def \showISBNx     #1{\unskip}     \fi
\ifx \showISBNxiii \undefined \def \showISBNxiii  #1{\unskip}     \fi
\ifx \showISSN     \undefined \def \showISSN      #1{\unskip}     \fi
\ifx \showLCCN     \undefined \def \showLCCN      #1{\unskip}     \fi
\ifx \shownote     \undefined \def \shownote      #1{#1}          \fi
\ifx \showarticletitle \undefined \def \showarticletitle #1{#1}   \fi
\ifx \showURL      \undefined \def \showURL       {\relax}        \fi
\providecommand\bibfield[2]{#2}
\providecommand\bibinfo[2]{#2}
\providecommand\natexlab[1]{#1}
\providecommand\showeprint[2][]{arXiv:#2}

\bibitem[Chang et~al\mbox{.}(2018)]%
        {chang2018rcaa}
\bibfield{author}{\bibinfo{person}{Xiaojun Chang}, \bibinfo{person}{PoYao Huang}, \bibinfo{person}{YiDong Shen}, \bibinfo{person}{Xiaodan Liang}, \bibinfo{person}{Yi Yang}, {and} \bibinfo{person}{Alexander~G Hauptmann}.} \bibinfo{year}{2018}\natexlab{}.
\newblock \showarticletitle{RCAA: Relational context-aware agents for person search}. In \bibinfo{booktitle}{\emph{Proceedings of the European Conference on Computer Vision (ECCV)}}. \bibinfo{pages}{84–100}.
\newblock


\bibitem[Chen et~al\mbox{.}(2018a)]%
        {chen2018ECCV}
\bibfield{author}{\bibinfo{person}{Dapeng Chen}, \bibinfo{person}{Hongsheng Li}, \bibinfo{person}{Xihui Liu}, \bibinfo{person}{Yantao Shen}, \bibinfo{person}{Jing Shao}, \bibinfo{person}{Zejian Yuan}, {and} \bibinfo{person}{Xiaogang Wang}.} \bibinfo{year}{2018}\natexlab{a}.
\newblock \showarticletitle{Improving Deep Visual Representation for Person Re-identification by Global and Local Image-language Association}. In \bibinfo{booktitle}{\emph{Proceedings of the European Conference on Computer Vision (ECCV)}}. \bibinfo{pages}{54--70}.
\newblock


\bibitem[Chen et~al\mbox{.}(2018c)]%
        {chen2018person}
\bibfield{author}{\bibinfo{person}{Di Chen}, \bibinfo{person}{Shanshan Zhang}, \bibinfo{person}{Wanli Ouyang}, \bibinfo{person}{Jian Yang}, {and} \bibinfo{person}{Ying Tai}.} \bibinfo{year}{2018}\natexlab{c}.
\newblock \showarticletitle{Person Search via a Mask-Guided Two-Stream CNN Model}. In \bibinfo{booktitle}{\emph{Proceedings of the European Conference on Computer Vision (ECCV)}}. \bibinfo{pages}{764--781}.
\newblock


\bibitem[{Chen} et~al\mbox{.}(2020b)]%
        {chen2020norm}
\bibfield{author}{\bibinfo{person}{Di {Chen}}, \bibinfo{person}{Shanshan {Zhang}}, \bibinfo{person}{Jian {Yang}}, {and} \bibinfo{person}{Bernt {Schiele}}.} \bibinfo{year}{2020}\natexlab{b}.
\newblock \showarticletitle{Norm-Aware Embedding for Efficient Person Search}. In \bibinfo{booktitle}{\emph{Proceedings of the Conference on Computer Vision and Pattern Recognition (CVPR)}}. \bibinfo{pages}{12615--12624}.
\newblock


\bibitem[Chen et~al\mbox{.}(2018b)]%
        {chen2018improving}
\bibfield{author}{\bibinfo{person}{Tianlang Chen}, \bibinfo{person}{Chenliang Xu}, {and} \bibinfo{person}{Jiebo Luo}.} \bibinfo{year}{2018}\natexlab{b}.
\newblock \showarticletitle{Improving text-based person search by spatial matching and adaptive threshold}. In \bibinfo{booktitle}{\emph{2018 IEEE Winter Conference on Applications of Computer Vision (WACV)}}. IEEE, \bibinfo{pages}{1879--1887}.
\newblock


\bibitem[{Chen} et~al\mbox{.}(2020a)]%
        {chen2020a}
\bibfield{author}{\bibinfo{person}{Xiaodong {Chen}}, \bibinfo{person}{Wu {Liu}}, \bibinfo{person}{Xinchen {Liu}}, \bibinfo{person}{Yongdong {Zhang}}, {and} \bibinfo{person}{Tao {Mei}}.} \bibinfo{year}{2020}\natexlab{a}.
\newblock \showarticletitle{A Cross-modality and Progressive Person Search System}. In \bibinfo{booktitle}{\emph{Proceedings of the 28th ACM International Conference on Multimedia}}. \bibinfo{pages}{4550--4552}.
\newblock


\bibitem[Chen et~al\mbox{.}(2015)]%
        {chen2015mirror}
\bibfield{author}{\bibinfo{person}{YingCong Chen}, \bibinfo{person}{WeiShi Zheng}, {and} \bibinfo{person}{Jianhuang Lai}.} \bibinfo{year}{2015}\natexlab{}.
\newblock \showarticletitle{Mirror Representation for Modeling View-Specific Transform in Person Re-Identification}. In \bibinfo{booktitle}{\emph{Proceedings of the 24th International Conference on Artificial Intelligence}}. \bibinfo{pages}{3402--3408}.
\newblock


\bibitem[{Dai} et~al\mbox{.}(2020)]%
        {dai2020dynamic}
\bibfield{author}{\bibinfo{person}{Ju {Dai}}, \bibinfo{person}{Pingping {Zhang}}, \bibinfo{person}{Huchuan {Lu}}, {and} \bibinfo{person}{Hongyu {Wang}}.} \bibinfo{year}{2020}\natexlab{}.
\newblock \showarticletitle{Dynamic imposter based online instance matching for person search}.
\newblock \bibinfo{journal}{\emph{Pattern Recognition}}  \bibinfo{volume}{100} (\bibinfo{year}{2020}), \bibinfo{pages}{107120}.
\newblock


\bibitem[Devlin et~al\mbox{.}(2018)]%
        {devlin2018bert}
\bibfield{author}{\bibinfo{person}{Jacob Devlin}, \bibinfo{person}{Ming-Wei Chang}, \bibinfo{person}{Kenton Lee}, {and} \bibinfo{person}{Kristina Toutanova}.} \bibinfo{year}{2018}\natexlab{}.
\newblock \showarticletitle{Bert: Pre-training of deep bidirectional transformers for language understanding}.
\newblock \bibinfo{journal}{\emph{arXiv preprint arXiv:1810.04805}} (\bibinfo{year}{2018}).
\newblock


\bibitem[{Dong} et~al\mbox{.}(2019)]%
        {dong2019person}
\bibfield{author}{\bibinfo{person}{Qi {Dong}}, \bibinfo{person}{Xiatian {Zhu}}, {and} \bibinfo{person}{Shaogang {Gong}}.} \bibinfo{year}{2019}\natexlab{}.
\newblock \showarticletitle{Person Search by Text Attribute Query As Zero-Shot Learning}. In \bibinfo{booktitle}{\emph{Proceedings of the IEEE international conference on computer vision (ICCV)}}. \bibinfo{pages}{3652--3661}.
\newblock


\bibitem[{Dong} et~al\mbox{.}(2020)]%
        {dong2020bi}
\bibfield{author}{\bibinfo{person}{Wenkai {Dong}}, \bibinfo{person}{Zhaoxiang {Zhang}}, \bibinfo{person}{Chunfeng {Song}}, {and} \bibinfo{person}{Tieniu {Tan}}.} \bibinfo{year}{2020}\natexlab{}.
\newblock \showarticletitle{Bi-Directional Interaction Network for Person Search}. In \bibinfo{booktitle}{\emph{Proceedings of the Conference on Computer Vision and Pattern Recognition (CVPR)}}. \bibinfo{pages}{2839--2848}.
\newblock


\bibitem[Fan et~al\mbox{.}(2020)]%
        {fan2020FGN}
\bibfield{author}{\bibinfo{person}{Zhibo Fan}, \bibinfo{person}{JinGang Yu}, \bibinfo{person}{Zhihao Liang}, \bibinfo{person}{Jiarong Ou}, \bibinfo{person}{Changxin Gao}, \bibinfo{person}{GuiSong Xia}, {and} \bibinfo{person}{Yuanqing Li}.} \bibinfo{year}{2020}\natexlab{}.
\newblock \showarticletitle{FGN: Fully Guided Network for Few-Shot Instance Segmentation}. In \bibinfo{booktitle}{\emph{Proceedings of the IEEE/CVF Conference on Computer Vision and Pattern Recognition}}. \bibinfo{pages}{9172--9181}.
\newblock


\bibitem[Gao et~al\mbox{.}(2021)]%
        {gao2021contextual}
\bibfield{author}{\bibinfo{person}{Chenyang Gao}, \bibinfo{person}{Guanyu Cai}, \bibinfo{person}{Xinyang Jiang}, \bibinfo{person}{Feng Zheng}, \bibinfo{person}{Jun Zhang}, \bibinfo{person}{Yifei Gong}, \bibinfo{person}{Pai Peng}, \bibinfo{person}{Xiaowei Guo}, {and} \bibinfo{person}{Xing Sun}.} \bibinfo{year}{2021}\natexlab{}.
\newblock \showarticletitle{Contextual Non-Local Alignment over Full-Scale Representation for Text-Based Person Search}.
\newblock \bibinfo{journal}{\emph{arXiv preprint arXiv:2101.03036}} (\bibinfo{year}{2021}).
\newblock


\bibitem[Gao et~al\mbox{.}(2020)]%
        {gao2020pose}
\bibfield{author}{\bibinfo{person}{Shang Gao}, \bibinfo{person}{Jingya Wang}, \bibinfo{person}{Huchuan Lu}, {and} \bibinfo{person}{Zimo Liu}.} \bibinfo{year}{2020}\natexlab{}.
\newblock \showarticletitle{Pose-guided visible part matching for occluded person ReID}. In \bibinfo{booktitle}{\emph{Proceedings of the IEEE/CVF Conference on Computer Vision and Pattern Recognition}}. \bibinfo{pages}{11744--11752}.
\newblock


\bibitem[Han et~al\mbox{.}(2019)]%
        {han2019re}
\bibfield{author}{\bibinfo{person}{Chuchu Han}, \bibinfo{person}{Jiacheng Ye}, \bibinfo{person}{Yunshan Zhong}, \bibinfo{person}{Xin Tan}, \bibinfo{person}{Chi Zhang}, \bibinfo{person}{Changxin Gao}, {and} \bibinfo{person}{Nong Sang}.} \bibinfo{year}{2019}\natexlab{}.
\newblock \showarticletitle{Re-id driven localization refinement for person search}. In \bibinfo{booktitle}{\emph{Proceedings of the IEEE/CVF International Conference on Computer Vision}}. \bibinfo{pages}{9814--9823}.
\newblock


\bibitem[{He} and {Zhang}(2018)]%
        {he2018end}
\bibfield{author}{\bibinfo{person}{Zhenwei {He}} {and} \bibinfo{person}{Lei {Zhang}}.} \bibinfo{year}{2018}\natexlab{}.
\newblock \showarticletitle{End-to-End Detection and Re-identification Integrated Net for Person Search}. In \bibinfo{booktitle}{\emph{Asian Conference on Computer Vision}}. \bibinfo{pages}{349--364}.
\newblock


\bibitem[{Hu} et~al\mbox{.}(2018)]%
        {hu2018squeeze}
\bibfield{author}{\bibinfo{person}{Jie {Hu}}, \bibinfo{person}{Li {Shen}}, {and} \bibinfo{person}{Gang {Sun}}.} \bibinfo{year}{2018}\natexlab{}.
\newblock \showarticletitle{Squeeze-and-Excitation Networks}. In \bibinfo{booktitle}{\emph{Proceedings of the IEEE/CVF Conference on Computer Vision and Pattern Recognition (CVPR)}}. \bibinfo{pages}{7132--7141}.
\newblock


\bibitem[{Islam}(2020)]%
        {islam2020person}
\bibfield{author}{\bibinfo{person}{Khawar {Islam}}.} \bibinfo{year}{2020}\natexlab{}.
\newblock \showarticletitle{Person search: New paradigm of person re-identification: A survey and outlook of recent works}.
\newblock \bibinfo{journal}{\emph{Image and Vision Computing}}  \bibinfo{volume}{101} (\bibinfo{year}{2020}), \bibinfo{pages}{103970}.
\newblock


\bibitem[{Ji} et~al\mbox{.}(2018)]%
        {ji2018fusion}
\bibfield{author}{\bibinfo{person}{Zhong {Ji}}, \bibinfo{person}{Shengjia {Li}}, {and} \bibinfo{person}{Yanwei {Pang}}.} \bibinfo{year}{2018}\natexlab{}.
\newblock \showarticletitle{Fusion-Attention Network for person search with free-form natural language}.
\newblock \bibinfo{journal}{\emph{Pattern Recognition Letters}}  \bibinfo{volume}{116} (\bibinfo{year}{2018}), \bibinfo{pages}{205--211}.
\newblock


\bibitem[Jiao et~al\mbox{.}(2021)]%
        {jiao2021instance}
\bibfield{author}{\bibinfo{person}{Bingliang Jiao}, \bibinfo{person}{Xin Tan}, \bibinfo{person}{Lu Yang}, \bibinfo{person}{Yunlong Wang}, {and} \bibinfo{person}{Peng Wang}.} \bibinfo{year}{2021}\natexlab{}.
\newblock \showarticletitle{Instance and Pair-Aware Dynamic Networks for Re-Identification}.
\newblock \bibinfo{journal}{\emph{arXiv preprint arXiv:2103.05395}} (\bibinfo{year}{2021}).
\newblock


\bibitem[Jing et~al\mbox{.}(2020)]%
        {jing2020pose}
\bibfield{author}{\bibinfo{person}{Ya Jing}, \bibinfo{person}{Chenyang Si}, \bibinfo{person}{Junbo Wang}, \bibinfo{person}{Wei Wang}, \bibinfo{person}{Liang Wang}, {and} \bibinfo{person}{Tieniu Tan}.} \bibinfo{year}{2020}\natexlab{}.
\newblock \showarticletitle{Pose-guided multi-granularity attention network for text-based person search}. In \bibinfo{booktitle}{\emph{Proceedings of the AAAI Conference on Artificial Intelligence}}, Vol.~\bibinfo{volume}{34}. \bibinfo{pages}{11189--11196}.
\newblock


\bibitem[{Jing} et~al\mbox{.}(2020)]%
        {jing2020cross}
\bibfield{author}{\bibinfo{person}{Ya {Jing}}, \bibinfo{person}{Wei {Wang}}, \bibinfo{person}{Liang {Wang}}, {and} \bibinfo{person}{Tieniu {Tan}}.} \bibinfo{year}{2020}\natexlab{}.
\newblock \showarticletitle{Cross-Modal Cross-Domain Moment Alignment Network for Person Search}. In \bibinfo{booktitle}{\emph{Proceedings of the Conference on Computer Vision and Pattern Recognition (CVPR)}}. \bibinfo{pages}{10678--10686}.
\newblock


\bibitem[{Krishna} et~al\mbox{.}(2017)]%
        {krishna2017visual}
\bibfield{author}{\bibinfo{person}{Ranjay {Krishna}}, \bibinfo{person}{Yuke {Zhu}}, \bibinfo{person}{Oliver {Groth}}, \bibinfo{person}{Justin {Johnson}}, \bibinfo{person}{Kenji {Hata}}, \bibinfo{person}{Joshua {Kravitz}}, \bibinfo{person}{Stephanie {Chen}}, \bibinfo{person}{Yannis {Kalantidis}}, \bibinfo{person}{LiJia {Li}}, \bibinfo{person}{David~A. {Shamma}}, \bibinfo{person}{Michael~S. {Bernstein}}, {and} \bibinfo{person}{Li {Fei-Fei}}.} \bibinfo{year}{2017}\natexlab{}.
\newblock \showarticletitle{Visual Genome: Connecting Language and Vision Using Crowdsourced Dense Image Annotations}.
\newblock \bibinfo{journal}{\emph{International Journal of Computer Vision}} \bibinfo{volume}{123}, \bibinfo{number}{1} (\bibinfo{year}{2017}), \bibinfo{pages}{32--73}.
\newblock


\bibitem[{Lan} et~al\mbox{.}(2018)]%
        {lan2018person}
\bibfield{author}{\bibinfo{person}{Xu {Lan}}, \bibinfo{person}{Xiatian {Zhu}}, {and} \bibinfo{person}{Shaogang {Gong}}.} \bibinfo{year}{2018}\natexlab{}.
\newblock \showarticletitle{Person Search by Multi-Scale Matching}. In \bibinfo{booktitle}{\emph{Proceedings of the European Conference on Computer Vision (ECCV)}}. \bibinfo{pages}{553--569}.
\newblock


\bibitem[Li et~al\mbox{.}(2017a)]%
        {li2017identity}
\bibfield{author}{\bibinfo{person}{Shuang Li}, \bibinfo{person}{Tong Xiao}, \bibinfo{person}{Hongsheng Li}, \bibinfo{person}{Wei Yang}, {and} \bibinfo{person}{Xiaogang Wang}.} \bibinfo{year}{2017}\natexlab{a}.
\newblock \showarticletitle{Identity-aware textual-visual matching with latent co-attention}. In \bibinfo{booktitle}{\emph{Proceedings of the IEEE International Conference on Computer Vision}}. \bibinfo{pages}{1890--1899}.
\newblock


\bibitem[Li et~al\mbox{.}(2017b)]%
        {li2017person}
\bibfield{author}{\bibinfo{person}{Shuang Li}, \bibinfo{person}{Tong Xiao}, \bibinfo{person}{Hongsheng Li}, \bibinfo{person}{Bolei Zhou}, \bibinfo{person}{Dayu Yue}, {and} \bibinfo{person}{Xiaogang Wang}.} \bibinfo{year}{2017}\natexlab{b}.
\newblock \showarticletitle{Person search with natural language description}. In \bibinfo{booktitle}{\emph{Proceedings of the IEEE Conference on Computer Vision and Pattern Recognition}}. \bibinfo{pages}{1970--1979}.
\newblock


\bibitem[Li et~al\mbox{.}(2016)]%
        {li2016null}
\bibfield{author}{\bibinfo{person}{Zhang Li}, \bibinfo{person}{Tao Xiang}, {and} \bibinfo{person}{Shaogang Gong}.} \bibinfo{year}{2016}\natexlab{}.
\newblock \showarticletitle{Learning a Discriminative Null Space for Person Re-Identification}. In \bibinfo{booktitle}{\emph{Proceedings of the IEEE Conference on Computer Vision and Pattern Recognition}}. \bibinfo{pages}{1239--1248}.
\newblock


\bibitem[{Lin} et~al\mbox{.}(2014)]%
        {lin2014microsoft}
\bibfield{author}{\bibinfo{person}{Tsung-Yi {Lin}}, \bibinfo{person}{Michael {Maire}}, \bibinfo{person}{Serge~J. {Belongie}}, \bibinfo{person}{James {Hays}}, \bibinfo{person}{Pietro {Perona}}, \bibinfo{person}{Deva {Ramanan}}, \bibinfo{person}{Piotr {Dollár}}, {and} \bibinfo{person}{C.~Lawrence {Zitnick}}.} \bibinfo{year}{2014}\natexlab{}.
\newblock \showarticletitle{Microsoft COCO: Common Objects in Context}. In \bibinfo{booktitle}{\emph{European Conference on Computer Vision}}. \bibinfo{pages}{740--755}.
\newblock


\bibitem[Liu et~al\mbox{.}(2017)]%
        {liu2017machines}
\bibfield{author}{\bibinfo{person}{Hao Liu}, \bibinfo{person}{Jiashi Feng}, \bibinfo{person}{Zequn Jie}, \bibinfo{person}{Karlekar Jayashree}, \bibinfo{person}{Bo Zhao}, \bibinfo{person}{Meibin Qi}, \bibinfo{person}{Jianguo Jiang}, {and} \bibinfo{person}{Shuicheng Yan}.} \bibinfo{year}{2017}\natexlab{}.
\newblock \showarticletitle{Neural person search machines}. In \bibinfo{booktitle}{\emph{Proceedings of the IEEE international conference on computer vision}}. \bibinfo{pages}{493–501}.
\newblock


\bibitem[Liu et~al\mbox{.}(2018)]%
        {liu2018pose}
\bibfield{author}{\bibinfo{person}{Jinxian Liu}, \bibinfo{person}{Bingbing Ni}, \bibinfo{person}{Yichao Yan}, \bibinfo{person}{Peng Zhou}, \bibinfo{person}{Shuo Cheng}, {and} \bibinfo{person}{Jianguo Hu}.} \bibinfo{year}{2018}\natexlab{}.
\newblock \showarticletitle{Pose transferrable person re-identification}. In \bibinfo{booktitle}{\emph{Proceedings of the IEEE Conference on Computer Vision and Pattern Recognition}}. \bibinfo{pages}{4099--4108}.
\newblock


\bibitem[{Liu} et~al\mbox{.}(2019)]%
        {liu2019deep}
\bibfield{author}{\bibinfo{person}{Jiawei {Liu}}, \bibinfo{person}{ZhengJun {Zha}}, \bibinfo{person}{Richang {Hong}}, \bibinfo{person}{Meng {Wang}}, {and} \bibinfo{person}{Yongdong {Zhang}}.} \bibinfo{year}{2019}\natexlab{}.
\newblock \showarticletitle{Deep Adversarial Graph Attention Convolution Network for Text-Based Person Search}. In \bibinfo{booktitle}{\emph{Proceedings of the 27th ACM International Conference on Multimedia}}. \bibinfo{pages}{665--673}.
\newblock


\bibitem[Martinel et~al\mbox{.}(2016)]%
        {Martinel2016reid}
\bibfield{author}{\bibinfo{person}{Martinel}, \bibinfo{person}{Niki}, \bibinfo{person}{Abir Das}, \bibinfo{person}{Christian Micheloni}, {and} \bibinfo{person}{Amit~K. Roy-Chowdhury}.} \bibinfo{year}{2016}\natexlab{}.
\newblock \showarticletitle{Temporal Model Adaptation for Person Re-Identification}. In \bibinfo{booktitle}{\emph{Proceedings of the European Conference on Computer Vision (ECCV)}}. \bibinfo{pages}{858–--877}.
\newblock


\bibitem[Munjal et~al\mbox{.}(2019)]%
        {munjal2019query}
\bibfield{author}{\bibinfo{person}{Bharti Munjal}, \bibinfo{person}{Sikandar Amin}, \bibinfo{person}{Federico Tombari}, {and} \bibinfo{person}{Fabio Galasso}.} \bibinfo{year}{2019}\natexlab{}.
\newblock \showarticletitle{Query-Guided End-To-End Person Search}. In \bibinfo{booktitle}{\emph{Proceedings of the IEEE/CVF Conference on Computer Vision and Pattern Recognition}}. \bibinfo{pages}{811--820}.
\newblock


\bibitem[{Munjal} et~al\mbox{.}(2019)]%
        {munjal2019knowledge}
\bibfield{author}{\bibinfo{person}{Bharti {Munjal}}, \bibinfo{person}{Fabio {Galasso}}, {and} \bibinfo{person}{Sikandar {Amin}}.} \bibinfo{year}{2019}\natexlab{}.
\newblock \showarticletitle{Knowledge Distillation for End-to-End Person Search.}. In \bibinfo{booktitle}{\emph{BMVC}}. \bibinfo{pages}{216}.
\newblock


\bibitem[Niu et~al\mbox{.}(2020)]%
        {niu2020improving}
\bibfield{author}{\bibinfo{person}{Kai Niu}, \bibinfo{person}{Yan Huang}, \bibinfo{person}{Wanli Ouyang}, {and} \bibinfo{person}{Liang Wang}.} \bibinfo{year}{2020}\natexlab{}.
\newblock \showarticletitle{Improving description-based person re-identification by multi-granularity image-text alignments}.
\newblock \bibinfo{journal}{\emph{IEEE Transactions on Image Processing}}  \bibinfo{volume}{29} (\bibinfo{year}{2020}), \bibinfo{pages}{5542--5556}.
\newblock


\bibitem[Sarafianos et~al\mbox{.}(2019)]%
        {sarafianos2019adversarial}
\bibfield{author}{\bibinfo{person}{Nikolaos Sarafianos}, \bibinfo{person}{Xiang Xu}, {and} \bibinfo{person}{Ioannis~A Kakadiaris}.} \bibinfo{year}{2019}\natexlab{}.
\newblock \showarticletitle{Adversarial representation learning for text-to-image matching}. In \bibinfo{booktitle}{\emph{Proceedings of the IEEE/CVF International Conference on Computer Vision}}. \bibinfo{pages}{5814--5824}.
\newblock


\bibitem[Sarfraz et~al\mbox{.}(2018)]%
        {sarfraz2018pose}
\bibfield{author}{\bibinfo{person}{M~Saquib Sarfraz}, \bibinfo{person}{Arne Schumann}, \bibinfo{person}{Andreas Eberle}, {and} \bibinfo{person}{Rainer Stiefelhagen}.} \bibinfo{year}{2018}\natexlab{}.
\newblock \showarticletitle{A pose-sensitive embedding for person re-identification with expanded cross neighborhood re-ranking}. In \bibinfo{booktitle}{\emph{Proceedings of the IEEE Conference on Computer Vision and Pattern Recognition}}. \bibinfo{pages}{420--429}.
\newblock


\bibitem[Suh et~al\mbox{.}(2018)]%
        {suh2018part}
\bibfield{author}{\bibinfo{person}{Yumin Suh}, \bibinfo{person}{Jingdong Wang}, \bibinfo{person}{Siyu Tang}, \bibinfo{person}{Tao Mei}, {and} \bibinfo{person}{Kyoung~Mu Lee}.} \bibinfo{year}{2018}\natexlab{}.
\newblock \showarticletitle{Part-aligned bilinear representations for person re-identification}. In \bibinfo{booktitle}{\emph{Proceedings of the European Conference on Computer Vision (ECCV)}}. \bibinfo{pages}{402--419}.
\newblock


\bibitem[Vu et~al\mbox{.}(2019)]%
        {vu2019cascade}
\bibfield{author}{\bibinfo{person}{Thang Vu}, \bibinfo{person}{Hyunjun Jang}, \bibinfo{person}{Trung~X Pham}, {and} \bibinfo{person}{Chang~Dong Yoo}.} \bibinfo{year}{2019}\natexlab{}.
\newblock \showarticletitle{Cascade RPN: Delving into High-Quality Region Proposal Network with Adaptive Convolution}. In \bibinfo{booktitle}{\emph{Conference on Neural Information Processing Systems (NeurIPS)}}. \bibinfo{pages}{1432--1442}.
\newblock


\bibitem[{Wang} et~al\mbox{.}(2020)]%
        {wang2020tcts}
\bibfield{author}{\bibinfo{person}{Cheng {Wang}}, \bibinfo{person}{Bingpeng {Ma}}, \bibinfo{person}{Hong {Chang}}, \bibinfo{person}{Shiguang {Shan}}, {and} \bibinfo{person}{Xilin {Chen}}.} \bibinfo{year}{2020}\natexlab{}.
\newblock \showarticletitle{TCTS: A Task-Consistent Two-Stage Framework for Person Search}. In \bibinfo{booktitle}{\emph{Proceedings of the IEEE Conference on Computer Vision and Pattern Recognition}}. \bibinfo{pages}{11952--11961}.
\newblock


\bibitem[Wang et~al\mbox{.}(2019a)]%
        {wang2019region}
\bibfield{author}{\bibinfo{person}{Jiaqi Wang}, \bibinfo{person}{Kai Chen}, \bibinfo{person}{Shuo Yang}, \bibinfo{person}{Chen~Change Loy}, {and} \bibinfo{person}{Dahua Lin}.} \bibinfo{year}{2019}\natexlab{a}.
\newblock \showarticletitle{Region Proposal by Guided Anchoring}. In \bibinfo{booktitle}{\emph{Proceedings of the IEEE/CVF Conference on Computer Vision and Pattern Recognition}}. \bibinfo{pages}{2965--2974}.
\newblock


\bibitem[Wang et~al\mbox{.}(2019b)]%
        {wang2019vehicle}
\bibfield{author}{\bibinfo{person}{Peng Wang}, \bibinfo{person}{Bingliang Jiao}, \bibinfo{person}{Lu Yang}, \bibinfo{person}{Yifei Yang}, \bibinfo{person}{Shizhou Zhang}, \bibinfo{person}{Wei Wei}, {and} \bibinfo{person}{Yanning Zhang}.} \bibinfo{year}{2019}\natexlab{b}.
\newblock \showarticletitle{Vehicle Re-Identification in Aerial Imagery: Dataset and Approach}. In \bibinfo{booktitle}{\emph{Proceedings of the IEEE international conference on computer vision}}. \bibinfo{pages}{460--469}.
\newblock


\bibitem[Xiao et~al\mbox{.}(2017)]%
        {xiao2017personsearch}
\bibfield{author}{\bibinfo{person}{Tong Xiao}, \bibinfo{person}{Shuang Li}, \bibinfo{person}{Bochao Wang}, \bibinfo{person}{Liang Lin}, {and} \bibinfo{person}{Xiaogang Wang}.} \bibinfo{year}{2017}\natexlab{}.
\newblock \showarticletitle{Joint detection and identification feature learning for person search}. In \bibinfo{booktitle}{\emph{Proceedings of the IEEE Conference on Computer Vision and Pattern Recognition}}. \bibinfo{pages}{3415–3424}.
\newblock


\bibitem[Xu et~al\mbox{.}(2014)]%
        {xu2014person}
\bibfield{author}{\bibinfo{person}{Yuanlu Xu}, \bibinfo{person}{Bingpeng Ma}, \bibinfo{person}{Rui Huang}, {and} \bibinfo{person}{Liang Lin}.} \bibinfo{year}{2014}\natexlab{}.
\newblock \showarticletitle{Person Search in a Scene by Jointly Modeling People Commonness and Person Uniqueness}. In \bibinfo{booktitle}{\emph{Proceedings of the 22nd ACM international conference on Multimedia}}. \bibinfo{pages}{937--940}.
\newblock


\bibitem[{Yan} et~al\mbox{.}(2019)]%
        {yan2019learning}
\bibfield{author}{\bibinfo{person}{Yichao {Yan}}, \bibinfo{person}{Qiang {Zhang}}, \bibinfo{person}{Bingbing {Ni}}, \bibinfo{person}{Wendong {Zhang}}, \bibinfo{person}{Minghao {Xu}}, {and} \bibinfo{person}{Xiaokang {Yang}}.} \bibinfo{year}{2019}\natexlab{}.
\newblock \showarticletitle{Learning Context Graph for Person Search}. In \bibinfo{booktitle}{\emph{Proceedings of the IEEE Conference on Computer Vision and Pattern Recognition}}. \bibinfo{pages}{2158--2167}.
\newblock


\bibitem[{Ye} et~al\mbox{.}(2021)]%
        {ye2021deep}
\bibfield{author}{\bibinfo{person}{Mang {Ye}}, \bibinfo{person}{Jianbing {Shen}}, \bibinfo{person}{Gaojie {Lin}}, \bibinfo{person}{Tao {Xiang}}, \bibinfo{person}{Ling {Shao}}, {and} \bibinfo{person}{Steven~C.H. {Hoi}}.} \bibinfo{year}{2021}\natexlab{}.
\newblock \showarticletitle{Deep Learning for Person Re-identification: A Survey and Outlook.}
\newblock \bibinfo{journal}{\emph{IEEE Transactions on Pattern Analysis and Machine Intelligence}} (\bibinfo{year}{2021}), \bibinfo{pages}{1--1}.
\newblock


\bibitem[{Zhai} et~al\mbox{.}(2019)]%
        {zhai2019fmt}
\bibfield{author}{\bibinfo{person}{Sulan {Zhai}}, \bibinfo{person}{Shunqiang {Liu}}, \bibinfo{person}{Xiao {Wang}}, {and} \bibinfo{person}{Jin {Tang}}.} \bibinfo{year}{2019}\natexlab{}.
\newblock \showarticletitle{FMT: fusing multi-task convolutional neural network for person search}.
\newblock \bibinfo{journal}{\emph{Multimedia Tools and Applications}} \bibinfo{volume}{78}, \bibinfo{number}{22} (\bibinfo{year}{2019}), \bibinfo{pages}{31605--31616}.
\newblock


\bibitem[Zhang et~al\mbox{.}(2021)]%
        {zhang2021person}
\bibfield{author}{\bibinfo{person}{Shizhou Zhang}, \bibinfo{person}{Qi Zhang}, \bibinfo{person}{Yifei Yang}, \bibinfo{person}{Xing Wei}, \bibinfo{person}{Peng Wang}, \bibinfo{person}{Bingliang Jiao}, {and} \bibinfo{person}{Yanning Zhang}.} \bibinfo{year}{2021}\natexlab{}.
\newblock \showarticletitle{Person Re-Identification in Aerial Imagery}.
\newblock \bibinfo{journal}{\emph{IEEE Transactions on Multimedia}}  \bibinfo{volume}{23} (\bibinfo{year}{2021}), \bibinfo{pages}{281--291}.
\newblock


\bibitem[{Zhang} et~al\mbox{.}(2020)]%
        {zhang2020diverse}
\bibfield{author}{\bibinfo{person}{Xinyu {Zhang}}, \bibinfo{person}{Xinlong {Wang}}, \bibinfo{person}{JiaWang {Bian}}, \bibinfo{person}{Chunhua {Shen}}, {and} \bibinfo{person}{Mingyu {You}}.} \bibinfo{year}{2020}\natexlab{}.
\newblock \showarticletitle{Diverse Knowledge Distillation for End-to-End Person Search}.
\newblock \bibinfo{journal}{\emph{arXiv preprint arXiv:2012.11187}} (\bibinfo{year}{2020}).
\newblock


\bibitem[Zhang and Lu(2018)]%
        {zhang2018deep}
\bibfield{author}{\bibinfo{person}{Ying Zhang} {and} \bibinfo{person}{Huchuan Lu}.} \bibinfo{year}{2018}\natexlab{}.
\newblock \showarticletitle{Deep cross-modal projection learning for image-text matching}. In \bibinfo{booktitle}{\emph{Proceedings of the European Conference on Computer Vision (ECCV)}}. \bibinfo{pages}{686--701}.
\newblock


\bibitem[{Zheng} et~al\mbox{.}(2020)]%
        {zheng2020segmentation}
\bibfield{author}{\bibinfo{person}{Dingyuan {Zheng}}, \bibinfo{person}{Jimin {Xiao}}, \bibinfo{person}{Kaizhu {Huang}}, {and} \bibinfo{person}{Yao {Zhao}}.} \bibinfo{year}{2020}\natexlab{}.
\newblock \showarticletitle{Segmentation mask guided end-to-end person search}.
\newblock \bibinfo{journal}{\emph{Signal Processing-image Communication}}  \bibinfo{volume}{86} (\bibinfo{year}{2020}), \bibinfo{pages}{115876}.
\newblock


\bibitem[Zheng et~al\mbox{.}(2015)]%
        {zheng2015scalable}
\bibfield{author}{\bibinfo{person}{Liang Zheng}, \bibinfo{person}{Liyue Shen}, \bibinfo{person}{Lu Tian}, \bibinfo{person}{Shengjin Wang}, \bibinfo{person}{Jingdong Wang}, {and} \bibinfo{person}{Qi Tian}.} \bibinfo{year}{2015}\natexlab{}.
\newblock \showarticletitle{Scalable person re-identification: A benchmark}. In \bibinfo{booktitle}{\emph{Proceedings of the IEEE international conference on computer vision}}. \bibinfo{pages}{1116--1124}.
\newblock


\bibitem[Zheng et~al\mbox{.}(2017a)]%
        {zheng2017PRW}
\bibfield{author}{\bibinfo{person}{Liang Zheng}, \bibinfo{person}{Hengheng Zhang}, \bibinfo{person}{Shaoyan Sun}, \bibinfo{person}{Yi Chandrakerand~Yang}, {and} \bibinfo{person}{Qi Tian}.} \bibinfo{year}{2017}\natexlab{a}.
\newblock \showarticletitle{Person re-identification in the wild}. In \bibinfo{booktitle}{\emph{Proceedings of the IEEE Conference on Computer Vision and Pattern Recognition}}. \bibinfo{pages}{1367–1376}.
\newblock


\bibitem[Zheng et~al\mbox{.}(2011)]%
        {zheng2011person}
\bibfield{author}{\bibinfo{person}{WeiShi Zheng}, \bibinfo{person}{Shaogang Gong}, {and} \bibinfo{person}{Tao Xiang}.} \bibinfo{year}{2011}\natexlab{}.
\newblock \showarticletitle{Person re-identification by probabilistic relative distance comparison}. In \bibinfo{booktitle}{\emph{Proceedings of the IEEE Conference on Computer Vision and Pattern Recognition}}. IEEE, \bibinfo{pages}{649--656}.
\newblock


\bibitem[Zheng et~al\mbox{.}(2017b)]%
        {zheng2017dual}
\bibfield{author}{\bibinfo{person}{Zhedong Zheng}, \bibinfo{person}{Liang Zheng}, \bibinfo{person}{Michael Garrett}, \bibinfo{person}{Yi Yang}, {and} \bibinfo{person}{YiDong Shen}.} \bibinfo{year}{2017}\natexlab{b}.
\newblock \showarticletitle{Dual-path convolutional image-text embedding with instance loss}.
\newblock \bibinfo{journal}{\emph{arXiv preprint arXiv:1711.05535}} (\bibinfo{year}{2017}).
\newblock


\bibitem[{Zhong} et~al\mbox{.}(2020)]%
        {zhong2020robust}
\bibfield{author}{\bibinfo{person}{Yingji {Zhong}}, \bibinfo{person}{Xiaoyu {Wang}}, {and} \bibinfo{person}{Shiliang {Zhang}}.} \bibinfo{year}{2020}\natexlab{}.
\newblock \showarticletitle{Robust Partial Matching for Person Search in the Wild}. In \bibinfo{booktitle}{\emph{Proceedings of the IEEE/CVF Conference on Computer Vision and Pattern Recognition (CVPR)}}. \bibinfo{pages}{6827--6835}.
\newblock


\bibitem[Zhou et~al\mbox{.}(2018)]%
        {zhou2018graph}
\bibfield{author}{\bibinfo{person}{Qin Zhou}, \bibinfo{person}{Heng Fan}, \bibinfo{person}{Shibao Zheng}, \bibinfo{person}{Hang Su}, \bibinfo{person}{Xinzhe Li}, \bibinfo{person}{Shuang Wu}, {and} \bibinfo{person}{Haibin Ling}.} \bibinfo{year}{2018}\natexlab{}.
\newblock \showarticletitle{Graph correspondence transfer for person re-identification}. In \bibinfo{booktitle}{\emph{Proceedings of the AAAI Conference on Artificial Intelligence}}, Vol.~\bibinfo{volume}{32}.
\newblock


\end{thebibliography}

\end{document}